\documentclass[10pt,journal,compsoc]{IEEEtran}
\usepackage[utf8]{inputenc}
\ifCLASSOPTIONcompsoc
\usepackage[nocompress]{cite}
\else
\usepackage{cite}
\fi

\ifCLASSINFOpdf
\else
\fi

\usepackage{xcolor}
\usepackage{color}
\usepackage{booktabs}
\usepackage{multirow}
\usepackage{colortbl}
\usepackage{graphicx}
\usepackage{caption}
\usepackage{array}
\usepackage{dblfloatfix}
\usepackage{balance}
\usepackage{enumitem}
\usepackage{url}
\setlist[enumerate]{itemsep=0mm}

\newcolumntype{C}[1]{>{\centering\arraybackslash}m{#1}}

\usepackage[framemethod=tikz]{mdframed}
\definecolor{mycolor}{rgb}{0.9, 0.0, 0.0}
\newmdenv[innerlinewidth=0.5pt, roundcorner=4pt,linecolor=mycolor,innerleftmargin=6pt,
innerrightmargin=6pt,innertopmargin=6pt,innerbottommargin=6pt]{rolebox}
\definecolor{mycolor1}{rgb}{0.0, 0.0, 0.9}
\newmdenv[innerlinewidth=0.5pt, roundcorner=4pt,linecolor=mycolor1,innerleftmargin=6pt,
innerrightmargin=6pt,innertopmargin=6pt,innerbottommargin=6pt]{commentbox}

\usepackage{bigstrut}
\usepackage{multirow}
\usepackage{amsmath}
\usepackage{subfigure}
\usepackage{chngcntr}

\makeatletter
\long\def\@IEEEtitleabstractindextextbox#1{\parbox{0.922\textwidth}{#1}}
\makeatother

\newcommand{\subparagraph}{}
\usepackage[compact]{titlesec}
\titlespacing{\section}{1pt}{*0}{*0}
\titlespacing{\subsection}{1pt}{*0}{*0}
\titlespacing{\subsubsection}{1pt}{*0}{*0}

\usepackage[rawfloats=true]{floatrow} 
\restylefloat{figure} 
\restylefloat{table} 

\newcommand{\Rev}[1]{\textcolor{black}{#1}}

\begin{document}
\bstctlcite{IEEEexample:BSTcontrol}

\title{Morphing Attack Detection - Database, Evaluation Platform and Benchmarking\thanks{The following paper is a pre-print. The article is accepted for publication in IEEE Transactions on Information Forensics and Security (TIFS).}}

\author{Kiran Raja$^*$, Matteo Ferrara$^\dagger$, Annalisa Franco$^\dagger$, Luuk Spreeuwers$^\ddagger$,   Ilias Batskos$^\ddagger$, Florens de Wit$^\ddagger$, \\Marta Gomez-Barrero$^{**}$,  Ulrich Scherhag$^{\ddagger\ddagger}$,  Daniel Fischer$^{\ddagger\ddagger}$, Sushma Venkatesh$^*$, Jag Mohan Singh$^*$, Guoqiang Li$^*$, Loïc Bergeron$^*$, Sergey Isadskiy$^{\ddagger\ddagger}$, Raghavendra Ramachandra$^*$, Christian Rathgeb$^{\ddagger\ddagger}$, Dinusha Frings$^\mathsection$, Uwe Seidel$^{\dagger\dagger}$, Fons Knopjes$^\mathsection$, Raymond Veldhuis$^\ddagger$, Davide Maltoni$^\dagger$, Christoph Busch$^*$\\
\normalsize{\textit{$^*$NTNU, Norway, $^\dagger$UBO, Italy, $^\ddagger$UTW,The Netherlands, $^{**}$HS-Ansbach, Germany, $^{\ddagger\ddagger}$HDA, Germany, \\ $^\mathsection$NOI, The Netherlands, $^{\dagger\dagger}$Bundeskriminalamt, Germany}}}

\markboth{Journal of \LaTeX\ Class Files,~Vol.~14, No.~8, August~2015}%
{Shell \MakeLowercase{\textit{et al.}}: Bare Demo of IEEEtran.cls for Computer Society Journals}

\IEEEtitleabstractindextext{%
\begin{abstract}
Morphing attacks have posed a severe threat to Face Recognition System (FRS). Despite the number of advancements reported in recent works, we note serious open issues such as \Rev{independent benchmarking, generalizability challenges and considerations to age, gender, ethnicity that are inadequately addressed}. Morphing Attack Detection (MAD) algorithms often are prone to generalization challenges as they are database dependent. The existing databases, mostly of semi-public nature, lack in diversity in terms of ethnicity, various morphing process and post-processing pipelines. Further, they do not reflect a realistic operational scenario for Automated Border Control (ABC) and do not provide a basis to test MAD on unseen data, in order to benchmark the robustness of algorithms. In this work, we present a new sequestered dataset for facilitating the advancements of MAD where the algorithms can be tested on unseen data in an effort to better generalize. The newly constructed dataset consists of facial images from 150 subjects from various ethnicities, age-groups and both genders. In order to challenge the existing MAD algorithms, the morphed images are with careful subject pre-selection created from the contributing images, and further post-processed to remove morphing artifacts. The images are also printed and scanned to remove all digital cues and to simulate a realistic challenge for MAD algorithms. Further, we present a new online evaluation platform to test algorithms on sequestered data. With the platform we can benchmark the morph detection performance and study the generalization ability. This work also presents a detailed analysis on various subsets of sequestered data and outlines open challenges for future directions in MAD research.
\end{abstract}

}

\maketitle

\IEEEdisplaynontitleabstractindextext

\IEEEpeerreviewmaketitle

\IEEEraisesectionheading{\section{Introduction}\label{sec:introduction}}
\IEEEPARstart{M}{orphing} attacks pose threats to Face Recognition Systems (FRS) by exploiting the tolerance towards intra-subject variations. Such attacks constitute a vulnerability in various applications like identity management, identity verified border crossing and visa management \cite{ferrara2014magic}. Morphing attacks consists of generating a composite image of two subjects resembling closely (for instance similar age and same ethnicity) and using the composite image to verify both the subject in an access control scenario. The composite image, hereafter referred as \textit{Morphed Image} should be of sufficient quality to obtain a score above the threshold recommended by a FRS in an automated face comparison system. It should also be of sufficiently high quality to fool a trained border guard when inspected manually\cite{ferrara2014magic}.

The morphed image can for instance be obtained by a malicious actor by colluding with a person having no criminal record to mask the identity of the malicious actor himself/herself, in order to obtain a new passport.
When a malicious actor is granted a valid identity document, he/she can use it for various purposes posing a risk to national security in the worst possible scenarios. With such an assertion, the initial work demonstrating the morphing attacks illustrated that commercial-off-the-shelf (COTS) FRS could be defeated with a given set of morphed images \cite{ferrara2014magic}. That study further assessed if  morphing attacks would succeed when presented to border guards. This means morphing attacks pose a threat to FRS systems and leave a major security risk to any nation where the malicious actor enters.
 
Initial studies have investigated various aspects of morphing attacks starting from analysing the vulnerability of FRS in detail \cite{raghavendra2016detecting, raghavendra2017transferable, scherhag2017vulnerability, gomez2017your} to providing measures to detect and mitigate the attacks effectively \cite{scherhag2019face, raghavendra2016detecting, damer2018morgan, damer2019multi, damer122019detect, damer2018detecting, scherhag2019detection, debiasi2018prnu}. Further, a number of works have focused on studying various parameters influencing the decisions of morphing attack detection subsystems, while other  works have focused on providing the set of metrics to gauge the strengths of Morphing Attack Detection (MAD) mechanisms. The works have also noted the vulnerability of FRS with respect to morphing attacks, when using the digital images and re-digitized images (digitally captured image which is printed and subsequently scanned/re-digitized). In pursuit of the current State Of The Art (SOTA) in MAD, we first review the related work in the next section.

\begin{table*}[htp]
	\centering
	 \resizebox{0.99\textwidth}{!}{
	\begin{tabular}{lccccc}
		\hline 
		& 									& Digital (D)/& Database &  & Mode \bigstrut\\
    	 Work & Morphing Method  									&Re-digitized(R) &  (\# Morphed images) & Detection Approach & (see Section~\ref{sec:classification-of-mad}) \bigstrut\\ 
    	 & & (Print-and-Scan)& & & \bigstrut\\ 
		\hline 
		\hline
		Ferrara et al. (2014) \cite{ferrara2014magic}*					&  GIMP GAP &  D 		& 12  		& - 					& -  \bigstrut\\ 
		\hline 
		Ferrara et al. (2016) \cite{Ferrara2016}*					&  GIMP GAP &  D 		& 21  		& - 					& - \bigstrut\\ 
		\hline
		Raghavendra et al. (2016) \cite{raghavendra2016detecting}*		&  GIMP GAP	&  D 		& 450 		&  Texture + Classifier & S-MAD\bigstrut\\ 
		\hline 
		Scherhag et al. (2017) \cite{scherhag2017vulnerability} 		&  GIMP GAP &  D \& R  	& 231		&  Texture + Classifier	& S-MAD \bigstrut\\ 
		\hline 
		Raghavendra et al. (2017) \cite{raghavendra2017face}*            & GIMP GAP  &  D \& R   & 1423 ($\times$2)        & Texture + Classifier & S-MAD\bigstrut\\ 
		\hline
		Raghavendra et al.(2017) \cite{raghavendra2017transferable}*	&  GIMP GAP &  D \& R	& 362		&  Deep-CNN				& S-MAD \bigstrut\\ 
		\hline 
		Gomez-Barrero et al. (2017) \cite{gomez2017your}*				& -			&  D		& 840		& -						& S-MAD \bigstrut\\ 
		\hline 
		Ferrara et al. (2018) \cite{Ferrara2018demorphing}                   & Sqirlz Morph 2.1 & D \& R &   100 &   Demorphing & D-MAD \bigstrut\\
		\hline
		Damer et al. (2018) \cite{damer2018morgan}						& GAN		&  D		& 1000		&  GAN Based Detection  & S-MAD \bigstrut\\ 
		\hline 
		Raghavendra et al. (2018) \cite{raghavendra2018detecting}      & GIMP-GAP  & D \& R     & 2518     & Color Space Texture + Classifier & S-MAD\bigstrut\\ 
		\hline
		Scherhag et al. (2019) \cite{scherhag2019detection}*            & OpenCV/dlib, & D \& R &   964 ($\times$3)   & PRNU + Classifier & S-MAD \bigstrut\\
		                                                        &  FaceFusion and FaceMorpher &  & &\bigstrut\\ 
		\hline
		Ferrara et al. (2019) \cite{ferrara2019face}					&  Sqirlz Morph 2.1 &  D \& R 		& 100  		& Deep Neural Networks & D-MAD \bigstrut\\ 
		\hline
		Ferrara et al. (2019) \cite{Ferrara2019Decoupling}*					&  Triangulation with Dlib-landmarks  &  D 		& 560 ($\times$36)  		& - 					& -  \bigstrut\\ 
		\hline
		Scherhag et al. (2020) \cite{scherhag2020deep}                   & OpenCV/dlib, FaceFusion,u & D \& R &   791+3246 ($\times$3) &   Deep Features & D-MAD \bigstrut\\ 
		&  FaceMorpher and UBO Morpher &  &  &\bigstrut\\ 					
			\hline
		Venkatesh et al. (2020) \cite{MorphStyleGAN2020}*                   & StyleGAN & D  &   2500  &   -  & S-MAD \bigstrut\\ 
		\bigstrut\\		
		\hline
	\end{tabular}
	}
	\vspace{-2mm}
\caption{State of the art in Morphing Attack Databases and Vulnerability Reporting (* indicates vulnerability demonstrated using COTS FRS)} 
\label{tab:related-works}
\end{table*}

\section{Related Work in Morphing Attacks on FRS and Databases}
Morphing attacks can be conducted in two specific types in a broader sense - (i) morphing attacks using digital images (ii) morphing attacks using re-digitized images (a.k.a. printed-and-scanned images). The former domain is inspired by the practices of various countries which allow to upload a digital representation of the face image for various applications such as passport renewal in UK \cite{ukpassport2019portal} and visa application in New Zealand\cite{nzvisa2019portal}. The latter is used in many countries where the passport/visa/identity-card applicant is requested to provide an image such as in India \cite{india2019portal} and in most European countries (e.g. in The Netherlands\cite{netherlands2019portal}) and this leaves the opportunity for a malicious actor to morph the facial image before it is printed. The image submitted by the applicant is thereafter re-digitized for digital processing and biometric enrolment. The earlier works have considered both scenarios and studied the impact of both types of attacks \cite{ferrara2014magic,raghavendra2017transferable, gomez2017your, scherhag2017vulnerability}. In this section, we review the key aspects of earlier works in both domains. While the literature is extensive in the recent years, we focus in this work to the most relevant works with new databases for MAD. The reader is further referred to Scherhag et al. \cite{scherhag2019face} for a detailed survey of the literature.

\subsection{Morphing Attacks Using Digital Images}
The first work illustrating morphing attacks was reported in 2014 by Ferrara et al. \cite{ferrara2014magic} where a set of morphed images was created using the AR Face Database \cite{martinez2000ar}. 5 pairs of images were morphed for male subjects and 5 pairs of female subjects for studying the vulnerability of FRS \cite{ferrara2014magic}.  Further, to supplement the study, one morphed image constituted by one male and one female subject and another morphed image constituted by 3 male subjects was employed. The studies specifically investigated the vulnerability of two commercial FRS - \Rev{Neurotechnology} VeriLook SDK 5.4 \cite{neurotechfrssdk} and Luxand SDK 4.0\cite{luxandfrssdk}. The initial studies asserted the success of all morphed images in reaching a match for both constituent subjects probe images and thereby illustrating the vulnerability of face recognition systems. In the following work by Raghavendra et al. \cite{raghavendra2016detecting}, the authors investigated the vulnerability on a larger set of grey scale images with 450 morphed samples from 110 different subjects on the \Rev{Neurotechnology} Verilook SDK\cite{neurotechfrssdk}. In the same work, the authors also proposed a first detection approach suitable for morphed images that are processed only in the digital domain. Further, Scherhag et al. \cite{scherhag2017vulnerability} conducted a similar analysis on using both a commercial SDK and OpenFace SDK - an open source face recognition SDK. In yet another work, Raghavendra et al. \cite{raghavendra2017transferable} employed a total of 431 morphed images to evaluate MAD mechanisms using deep neural networks. In a complementary work, Gomez-Barrero et al. \cite{gomez2017your} investigated the vulnerability of FRS to morphing attacks using 840 images from the Multimodal BioSecure Database \cite{ortega2009multiscenario} in the digital domain and also investigated the vulnerability of fingerprint and iris biometric systems against biometric attacks. As an alternative to morphing approaches, Raghavendra et al. \cite{raghavendra2017face} presented another concept of averaging facial images and proved the vulnerability of FRS for morphed and averaged images in the digital domain. The vulnerability was reported again using the \Rev{Neurotechnology} Verilook SDK on a newly created database of 580 morphed images and 580 averaged images. In a different paradigm, Damer et al. \cite{damer2018morgan} presented an approach of generating morphed images using Generative Adversarial Networks (GAN) on a set of 1500 images to create 1000 morphed images. The authors compared the results of MAD mechanism against traditional Landmark Aligned (LMA) morphing approaches, the vulnerability of the generated database was reported using two open source face SDKs based on VGG Network \cite{simonyan2014very} and OpenFace \cite{amos2016openface}. The database was used to devise MAD mechanisms on digital images alone in following works \cite{damer2019multi, damer122019detect, damer2018detecting, scherhag2019detection, debiasi2018prnu}. 

\subsection{Morphing Attacks Using Print and Scanned Images} 
Motivated by threats of morphed images to FRS, a number of works have also investigated morphing attacks using re-digitized images (printed and scanned). The key assertion behind these works is that the loss of pixel level information, which was originally introduced by the morphing process, and is now lost due to subsequent printing and scanning processes using devices of various vendors decreases the MAD capability. Further the printing and scanning processes cause additional noise artifacts contained in the re-digitized morphed images \cite{scherhag2017vulnerability,raghavendra2017face, raghavendra2018detecting, hildebrandt2017benchmarking, Ferrara2018demorphing} 

The works in detecting re-digitized images employ the same techniques to generate morphs and then print-and-scan them. Raghavendra et al.\cite{raghavendra2017face} introduced a print and scanned database of 1423 morphed images using both morphing and averaging of pixels. The images were printed using a RICOH MPC 6003 SP on high-quality photo paper with 300 $g/m^2$ density and scanned using a HP Photosmart 5520 scanner at 300 dpi for bona fide, morphed and averaged images. The work also illustrated the vulnerability of COTS FRS with regards to re-digitized images to be equal to digital domain images while the MAD performance dropped. The same work was further extended with a database to have 2518 morphed images \cite{raghavendra2018detecting}.  In a similar direction, Scherhag et al.\cite{scherhag2019detection}, introduced a printed-scanned morphed face image database generated using the FRGCv2 face dataset. The authors used the Epson DS-50000 Scanner at 300 dpi to print and scan the morphed images generated using three different morphing schemes (OpenCV/dlib, FaceFusion and FaceMorpher) \cite{scherhag2019detection}. Ferrara et al. \cite{Ferrara2018demorphing} also introduced a  printed-scanned database for MAD, specifically to study the demorphing approach where the authors subtract the re-digitized images to detect a face morphing attack. The morphed images were printed and scanned at 600 dpi using a professional quality photoprinter \cite{Ferrara2018demorphing}.

\begin{figure}[htp]
	\centering
	\includegraphics[width=0.98\textwidth]{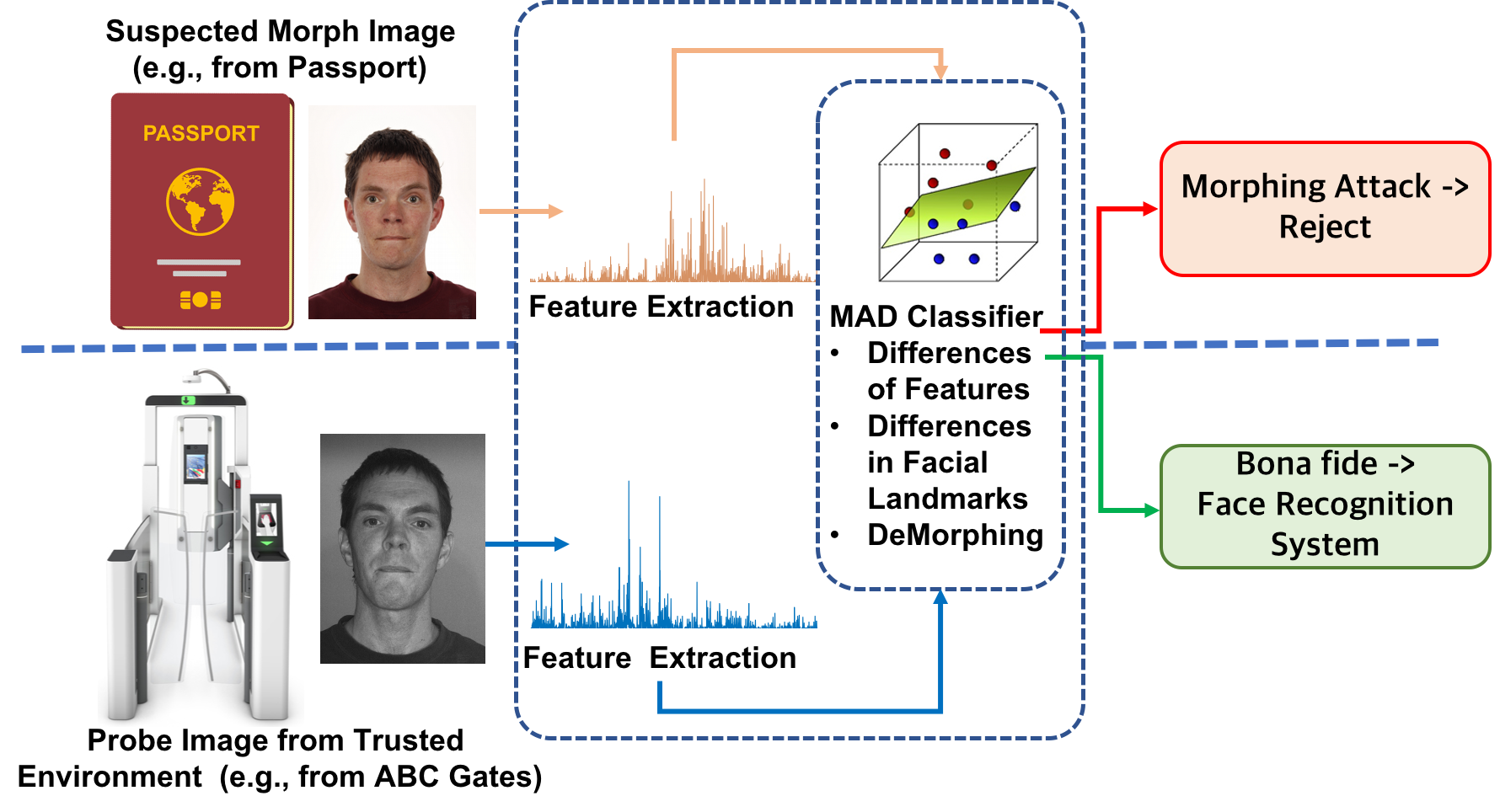}
	\caption{An illustration of the D-MAD pipeline}
	\label{fig:D-MAD-illustration}
\end{figure}

\subsection{Classification of MAD}
\label{sec:classification-of-mad}
While the aforementioned works have employed various databases, most of the works have also reported MAD mechanisms correspondingly to mitigate the threats on FRS: The algorithms for MAD can be classified in two classes:
\begin{itemize}[leftmargin=*]
    \item \textbf{Differential-image MAD (D-MAD)}: A suspected morph image is compared against an image captured in a trusted environment (e.g., ABC gate) to determine if the suspected image is morphed.
    \item \textbf{Single-image MAD (S-MAD)}: A suspected morph image is investigated (e.g. in a forensic process), in order to determine if the image itself is morphed without using any prior information or another reference image (captured under a trusted acquisition scenario).
\end{itemize}
We provide a brief review of the relevant algorithms reported in the recent works for both S-MAD and D-MAD.

\subsubsection{Differential-image MAD}
\label{sec:D-MAD-algorithms}
The general principle behind the D-MAD algorithms relies on the idea that given a suspected morphed image, $I_s$ and a reference image $I_t$ captured in a trusted environment, the difference between $I_s$ and $I_t$ is obtained. The lower the difference, either in the image space or feature space, the larger the probability that the suspected image is accepted as non-morphed (or bona fide image). The first approach of D-MAD was based on inverting the morphing process in a reverse engineered manner which was termed as \textit{Demorphing} \cite{Ferrara2018demorphing}. In a similar manner, a number of works have been reported where the difference of feature vectors from the bona fide image and from the morph image is used to determine if the suspected image is morphed \cite{scherhag2020deep,mohan2019robust}. The deep features from two different networks are employed to determine the difference in features in \cite{scherhag2020deep}, and features from the 3D shape and the diffuse reflectance component estimated directly from the image was employed to detect a morphing attack in \cite{mohan2019robust}. Another set of works explored the shift in landmarks of bona fide and suspected morph images in face region to determine the morphing attack \cite{damer2018detecting,scherhag2019detection}. For the sake of simplicity a generic illustration of the D-MAD working principle is presented in Figure~\ref{fig:D-MAD-illustration}.

\subsubsection{Single-image MAD}
\label{sec:S-MAD-algorithms}
S-MAD algorithms largely rely on learning a classifier to distinguish the bona fide image from a morphed image. Given a suspected morph image, $I_s$, the texture information is extracted from the normalized and aligned face. The texture features such as Binarized Statistical Image Features (BSIF) and Local Binary Patterns (LBP) are used to classify the images using a pre-trained SVM \Rev{classifier} \cite{raghavendra2018detecting,raghavendra2017face,scherhag2017vulnerability} in the earlier works. In a very similar direction, the LBP features were also explored in \cite{spreeuwers2018towards,scherhag2019detection}. While extending the works for MAD, another approach was proposed to exploit the colour spaces and the scale spaces jointly \cite{raghavendra2018detecting, ramachandra2019towards}. With the intent to address also the post-processed morphed images, pre-trained deep networks for extraction of texture features were employed to detect the morphing attacks not only in the digital domain, but also in re-digitized domain (print-scan) \cite{raghavendra2017transferable}. Notably, the earlier works have employed two deep neural networks including VGG19 \cite{simonyan2014very} and AlexNet \cite{krizhevsky2012imagenet}, where they perform feature level fusion of the first fully connected layers from both the networks \cite{raghavendra2017transferable}. In a continued effort, other deep networks have been investigated for detecting morph attacks \cite{ferrara2019face}. Another approach to detecting morphing attacks was proposed by extracting the features from the “Photo Response Non-Uniformity“ where the characteristics of the image sensor were employed to determine, if the image was morphed or not \cite{debiasi2018prnu}. Motivated by the effectiveness of the noise modelling, \Rev{better performing} algorithms have been reported where the color space has been investigated to seek for residuals of the morphing process \cite{venkatesh2019morphed} including dedicated context aggregation networks to automatically model the noise \cite{venkatesh2020detecting}.

\subsection{Limitations}
As noted from the set of works listed in the previous section and Table~\ref{tab:related-works}, there is a need for standardized and reproducible testing of MAD mechanisms. The limitations can be further divided in four main categories:

\begin{itemize}[leftmargin=*]
    \item \textbf{Need for cross-dataset evaluation:} As different works have used in-house datasets generated using different approaches, the proposed methods are only evaluated on limited sets. Despite the proposed MAD approaches performing very well on the in-house datasets, no works have attempted to study the generalizable detection performance except in recent works \cite{venkatesh2020detecting, spreeuwers2018towards} which attempts to study the cross-dataset evaluation. The missing aspect from different studies suffer from validation of SOTA proposed approaches in terms of generalizable detection performance and also indicating the directions for future works. In order to address this aspect, it is necessary to avoid the classical over-fitting problem for MAD mechanisms.
    \item \textbf{Need for sequestered database:} Further to support the reporting of generalizable detection performance in studies, there is a need for sequestered data for testing the robustness of the MAD algorithms. Thus, the need for a sequestered dataset, to which researchers do not have access for training purposes, is obvious. Sequestered data should solely be used for reproducible testing. Such tests on unknown data will establish a reliable benchmark of  algorithms and will indicate, whether said algorithms are robust to handle various factors unaware to researchers.
    \item \textbf{Need for independent evaluation:} As a third factor, MAD algorithms are often tuned to perform well on known datasets owing to the nature of in-house datasets. Despite the datasets being divided in training, testing and validation sets, it can be well observed that the algorithms and researchers have full access to look at the cases during an introspection and thereby improve their own MAD detection performance iteratively. While this enables continuous development and impovement of algorithms, morphing attacks in a real-life border crossing scenario can be compared to biometrics in the wild, where neither morphing generation algorithms, nor the post-processing approaches or printing and scanning mechanisms can be fully controlled. For the algorithms to be ready for operational deployment, there is a need for independent testing using morphed images which are unknown to the developers.  
    \item \textbf{Need for evaluation platform:} While independent testing is desired, there are not many organizations hosting such platforms limiting the researchers to devise robust algorithms. Although a similar evaluation effort is carried out by NIST \cite{NistFrvtMorph}, the NIST FRVT MORPH dataset, especially the subset containing post-processed print-scan and operational ABC gate images, is currently limited in size.
    Therefore, the need for an independent evaluation platform that runs continuously is needed to facilitate algorithmic evaluation and benchmark the detection performance against other competing algorithms in the lines of earlier evaluation platforms from University of Bologna, who have provided a long-standing fingerprint evaluation system \cite{dorizzi2009FVConGoing, FVConGoing}.
\end{itemize}

\subsection{Contributions of this work}
In order to address these four key limitations, in this work we provide three major contributions followed by the benchmarking of SOTA MAD mechanisms.
\begin{itemize}[leftmargin=*]
    \item A large scale sequestered database of morphed and bona fide images collected in three different sites constituting to	$1800$ photographs of $150$ subjects is released along with this article. The database covers various age groups, equal representation of genders and varied ethnicity making it an unique database for MAD algorithm evaluation.  The morphing of images was conducted with 6 different morphing algorithms presenting a wide variety of possible approaches. The images in the database consist of 5,748 morphed face images, where subsets consist of: (1) morphed images without post-processing to remove digital artifacts, (2) morphed and post-processed images to remove artifacts induced while morphing to produce passport quality ICAO photos \cite{ICAO-9303-p9-2015}, (3) printed and scanned versions of ICAO standard passport images using different combinations of printers and scanners including the scanners used in federal ID management offices in Europe. The database is accessible through the FVConGoing platform \cite{FVConGoing} to allow third parties for evaluation and benchmarking. 
    \item An unbiased and independent evaluation of $5$ state of the art MAD algorithms against 5,748 morphed face images and 1,396 bona fide face images. A total of 500,200 attempts with bona fide (69,800) and morphed (430,400) face images are evaluated to report the detection performance of current SOTA MAD mechanisms. 
    \item A new and independent evaluation platform is further presented to facilitate reproducible research where any researcher, governmental agency or private entity can upload SDKs and measure the performance of their MAD algorithm. The platform provides the benchmarking of the MAD performance against all previously submitted algorithms and specifically provides the results for different subsets corresponding to age, gender or ethnicity. Such detailed analysis will enable the researchers to identify the performance limitations of MAD mechanisms and facilitate them to develop more robust algorithms.
\end{itemize}

In the remainder of this article, in Section~\ref{sec:sotamd-database} we present the newly composed database where the details of the entire dataset are described. The new independent evaluation platform is introduced in Section~\ref{sec-evaluation-platform}. In Section~\ref{sec:mad-algoritms}, we present the set of SOTA algorithms that are particularly evaluated on the sequestered dataset. A detailed discussion of results and the analysis of MAD performance is reported in Section~\ref{sec:results-discussion}. While in Section~\ref{sec:conclusion} we draw the conclusions and list current limitations with the intention, to facilitate the efforts for development of future algorithms.

\begin{table}[!htbp]
  \centering
  \caption{Number of images in the database.}
  \resizebox{0.85\textwidth}{!}{
    \begin{tabular}{|l|c|c|}
    \hline
          & \multicolumn{1}{l|}{\cellcolor[rgb]{ .851,  .851,  .851}\textbf{Digital}} & \multicolumn{1}{l|}{\cellcolor[rgb]{ .851,  .851,  .851}\textbf{Printed\&Scanned}} \\
    \hline
    \rowcolor[rgb]{ .851,  .851,  .851} \textbf{Bona fide enrolment} & \cellcolor[rgb]{ 1,  1,  1}300 & \multicolumn{1}{c|}{\cellcolor[rgb]{ 1,  1,  1}1096} \\
    \hline
    \rowcolor[rgb]{ .851,  .851,  .851} \textbf{Morphed enrolment} & \cellcolor[rgb]{ 1,  1,  1}2045 & \multicolumn{1}{c|}{\cellcolor[rgb]{ 1,  1,  1}3703} \\
    \hline
    \rowcolor[rgb]{ .851,  .851,  .851} \textbf{Gate (Trusted live capture)} & \cellcolor[rgb]{ 1,  1,  1}1500 & \cellcolor[rgb]{ 1,  1,  1}- \\
    \hline
    \end{tabular}%
    }
  \label{tab:db_images}%
\end{table}%

\begin{table}[!htbp]
\centering
\caption{Minimum and maximum image size.}
\begin{tabular}{c|c|c|c|c|}
\cline{2-5}
 & \multicolumn{2}{c|}{\cellcolor[rgb]{ .851,  .851,  .851}\textbf{Enroll}} & \multicolumn{2}{c|}{\cellcolor[rgb]{ .851,  .851,  .851}\textbf{Gate}} \\ \cline{2-5} 
 & \cellcolor[rgb]{ .851,  .851,  .851}\textbf{Min} & \cellcolor[rgb]{ .851,  .851,  .851}\textbf{Max} & \cellcolor[rgb]{ .851,  .851,  .851}\textbf{Min} & \cellcolor[rgb]{ .851,  .851,  .851}\textbf{Max} \\ \hline
\multicolumn{1}{|c|}{\cellcolor[rgb]{ .851,  .851,  .851}\textbf{Original}} & 833x1111 & 5184x3456 & 383x533 & 1920x2560 \\ \hline
\multicolumn{1}{|c|}{\cellcolor[rgb]{ .851,  .851,  .851}\textbf{Digital}} & 362x482 & 4140x5323 & 381x508 & 997x1330 \\ \hline
\multicolumn{1}{|c|}{\cellcolor[rgb]{ .851,  .851,  .851}\textbf{P\&S}} & 337x449 & 552x709 & \multicolumn{2}{c|}{-} \\ \hline
\end{tabular}
\label{tab:db_imgSize}
\end{table}

\section{SOTAMD Database}
\label{sec:sotamd-database}
As noted in the earlier works, the existing MAD efforts by research institutions are largely based on internally created databases, which often are limited in size, diversity of image capture devices, image quality, realistic post-processing, and variability of morphing algorithms. We note that a best practice of using different databases and image acquisition and testing protocols makes it challenging, to benchmark MAD algorithms and thereby makes it for an operator next to impossible to judge the applicability of current MAD for operational deployment. In order to overcome these limitations and provide a new dataset for benchmarking (both for S-MAD and D-MAD algorithms) under realistic conditions with high quality images, we created a new dataset, to which we refer as State of the Art Morphing Detection (SOTAMD) dataset. The dataset consists of:
\begin{enumerate}[leftmargin=*]
  \item \textbf{Enrolment images}: bona fide face images taken in a capture set-up, which is meeting the requirements of passport application photo capture (e.g., photographer studio).
  \item \textbf{Gate images}: bona fide face images captured live with a face capture system in an Automated Border Control (ABC) gate.
  \item \textbf{Chip images}: compressed face images stored on an electronic Machine Readable Travel Document (e-MRTD).
  \item \textbf{Morphed face images}: morphed images created from the pool of passport face images. The database contains different kinds of morphed images as listed below:
  \begin{enumerate}
    \item \textbf{Digital morphed images}: Images obtained obtained directly after morphing in the digital domain.
    \item \textbf{Digital post-processed morphed images}: morphed images that are processed (automatically or manually) in the digital domain, to eliminate or hide the artifacts resulting from a morphing process.
    \item \textbf{Print-scanned morphed images}: post-processed morphed images that are printed and scanned to simulate the passport application process.
  \end{enumerate}
\end{enumerate}

A number of factors are considered in creating this dataset as a joint effort in an EU funded project - State-Of-The-Art-Morphing-Detection (SOTAMD) which are explained in the subsequent sections.\\
\Rev{Some information about the number of images in the database and their size is given, respectively, in Table \ref{tab:db_images} and Table \ref{tab:db_imgSize}. The bona fide enrolment images have been cropped to remove the background and resized in order to follow the same inter-eye distance distribution of the morphed images, so that it's not possible to infer the image class from its size. The details of the various subsets of data along with the details on morphing methods, print-scan pipeline, and compression details is provided in Table~\ref{tab:morphing-dataset-acronyms} and Table~\ref{tab:print-scan-details}.}. The images from the database are used to test both S-MAD and D-MAD algorithms according to the testing protocols defined in Section \ref{sec:evaluation-protocols}.

\begin{table}[!htbp]
  \centering
  \resizebox{0.85\textwidth}{!}{
    \begin{tabular}{|c|c|c|c|c|}
    \hline
    \rowcolor[rgb]{.851,.851,.851} \multicolumn{2}{|c|}{Gender} & \multicolumn{3}{c|}{Age} \bigstrut\\
    \hline
    Male & Female & A18-A35 & A36-A55 & A56-A75 \bigstrut\\
    \hline
    86 & 64 & 87 & 47 & 16 \bigstrut\\
    \hline
    \rowcolor[rgb]{.851,.851,.851} \multicolumn{5}{|c|}{Ethnicity} \bigstrut\\
    \hline
    European & African & India-Asian & East-Asian & Middle-Eastern \bigstrut\\
    \hline
    96 & 26 & 10 & 9 & 9 \bigstrut\\
    \hline
    \end{tabular}%
    }
    \caption{Demographics of the SOTAMD database}
    \label{tab:demographics}%
\end{table}%

\begin{table}[!htp]
  \centering
  \resizebox{0.8\textwidth}{!}{
    \begin{tabular}{|C{5em}|C{5em}|C{4.5em}|C{3em}|}
    \hline
    \rowcolor[rgb]{.851,.851,.851}  & Automated Morphing & {Manually post-processed} & Total\\
    \hline
    Digital images &  1475 & 570 & 2045 \\
    \hline
    Printed \& Scanned &  1453 & 2250 & 3703 \\
    \hline
    Total &  2928 & 2820 & 5748 \\
    \hline
    \end{tabular}
    }
  \caption{Total number of images with morphing and manual post-processing.}
  \label{tab:morphs}
\end{table}

\begin{table*}[b]
  \centering
  \resizebox{0.8\textwidth}{!}{
    \begin{tabular}{cccc}
    \hline
    \rowcolor[rgb]{.851,.851,.851} Partner  & Algorithm description & Automated & Manual  \bigstrut\\
     \rowcolor[rgb]{.851,.851,.851}  &  &  Post-processing method &  Post-processing method \bigstrut\\
    \hline
    Hochschule Darmstadt & FaceMorpher \cite{FaceMorpher}  & Facemorpher's internal &  No Manual \bigstrut\\
     &  &  post-processing +sharpening  & Post-processing \bigstrut\\
    \hline
    Hochschule Darmstadt & FaceFusion \cite{FaceFusion} & FaceFusion's internal & GIMP \bigstrut\\
      & (only used by HDA)  & post-processing+sharpening &   retouching\cite{GIMP} \bigstrut\\
    \hline
    Norwegian University of  & FaceMorph  &  The replacement of the eyeregion &  GIMP \bigstrut\\
     Science and Technology & (OpenCV with Dlib) \cite{dlib}  &  is performed in post-processing, &  retouching\cite{GIMP}  \bigstrut\\
    &   & to prevent a double iris. & \bigstrut\\
    \hline
    Norwegian University of &  FantaMorph \cite{Fantamorph}  &  Fantamorph's &  Adobe Photoshop \bigstrut\\
   Science and Technology &   (only used byNTN)  &  internal processing &  Retouching \cite{Photoshop} \bigstrut\\
    \hline
    University of Bologna & Triangulation with & Background replacement,edge & Adobe Photoshop \bigstrut\\
     &  Dlib-landmarks &  suppression, colour equalization & Retouching \cite{Photoshop} \bigstrut\\
    \hline
    University of Twente & Triangulation with &  Background replacement, & GIMP  \bigstrut\\
     &  STASM-landmarks \cite{STASM}  & Poisson image editing \cite{10.1145/1201775.882269}  &  retouching\cite{GIMP} \bigstrut\\
    \hline
    University of Bologna & Triangulation with & Background replacement  & GIMP  \bigstrut\\
     & NT-landmarks & edge suppression, colour equalization &  retouching\cite{GIMP} \bigstrut\\
    \hline
    \end{tabular}%
    }
    \caption{Contributed morphing methods, manual post-processing methods and automated post-processing methods.}
  \label{tab:morphing-dataset-methos-processing}%
\end{table*}%

\subsection{Subject Pre-selection}
An important aspect of creating a successful morph attack is subject selection, such that closely resembling pairs of faces are chosen \cite{scherhag2017vulnerability}. Following the guidelines of earlier works, the SOTAMD database was created by selecting the morph pairing candidates with high similarity with careful considerations to age, gender and ethnicity. As an additional measure, the selected morph pairing candidates were also validated by observing the comparison scores from two specific commercial-off-the-shelf (COTS) FRS - \Rev{Neurotechnology} Verilook SDK\cite{neurotechfrssdk} and Cognitec FaceVacs SDK \cite{cognitecfrssdk}. All the morphed images that did not verify against probe images from both contributing subjects were classified as \textit{low quality morph set} in the final database. This labeling makes the SOTAMD database highly relevant to investigate low quality and high quality morph detection capability. Such elimination and careful selection has led to 75 unique pairs of candidates for morphing from a total of 150 individuals of various ethnicity and age group. The subjects were selected amongst university staff and student corpus, and a casting agency website. Table \ref{tab:demographics} presents the gender, age and ethnicity demographics of the selected subjects for the final SOTAMD database.

\subsection{Bona Fide enrolment images}
For each of the 150 subjects in the SOTAMD database, two enrolment images were captured in high quality studio acquisition set-up reflecting the real-life passport photo capture process. Further, the enrolment images are also printed and scanned to have both digital and correspondingly printed and scanned subsets. The print and scan processes are conducted using various printers and scanners to increase the diversity of the dataset.\\
Given the nature of this work reflecting a operational border control scenario, we have exercised care to make sure the images are ICAO complaint \cite{ICAO-9303-p9-2015}. Thus, each of the images in the enrolment set was processed with professional software to comply with ICAO standards for eMRTD images. The processed images were further used for printing and scanning to closely follow the actual production scenario of passports based on the regulations in the Netherlands and Germany under EU member state regulations.  \\
The number of bona fide enrolment images in the new SOTAMD database is 300 in digital format, and 1096 printed and scanned. 
\subsection{Morphed enrolment images}
To simulate the criminal attack, we generated a number of morphed images to be used for enrolment, i.e. to be hypothetically presented \Rev{during the passport application process}. The morphed images have been created starting from the bona fide enrolment images (one for each subject).

\begin{figure}[!htp]
\begin{center}
\subcapcentertrue
\subfigure[Bona fide face image (Contributor A)]{
    \includegraphics[width=0.2\linewidth]{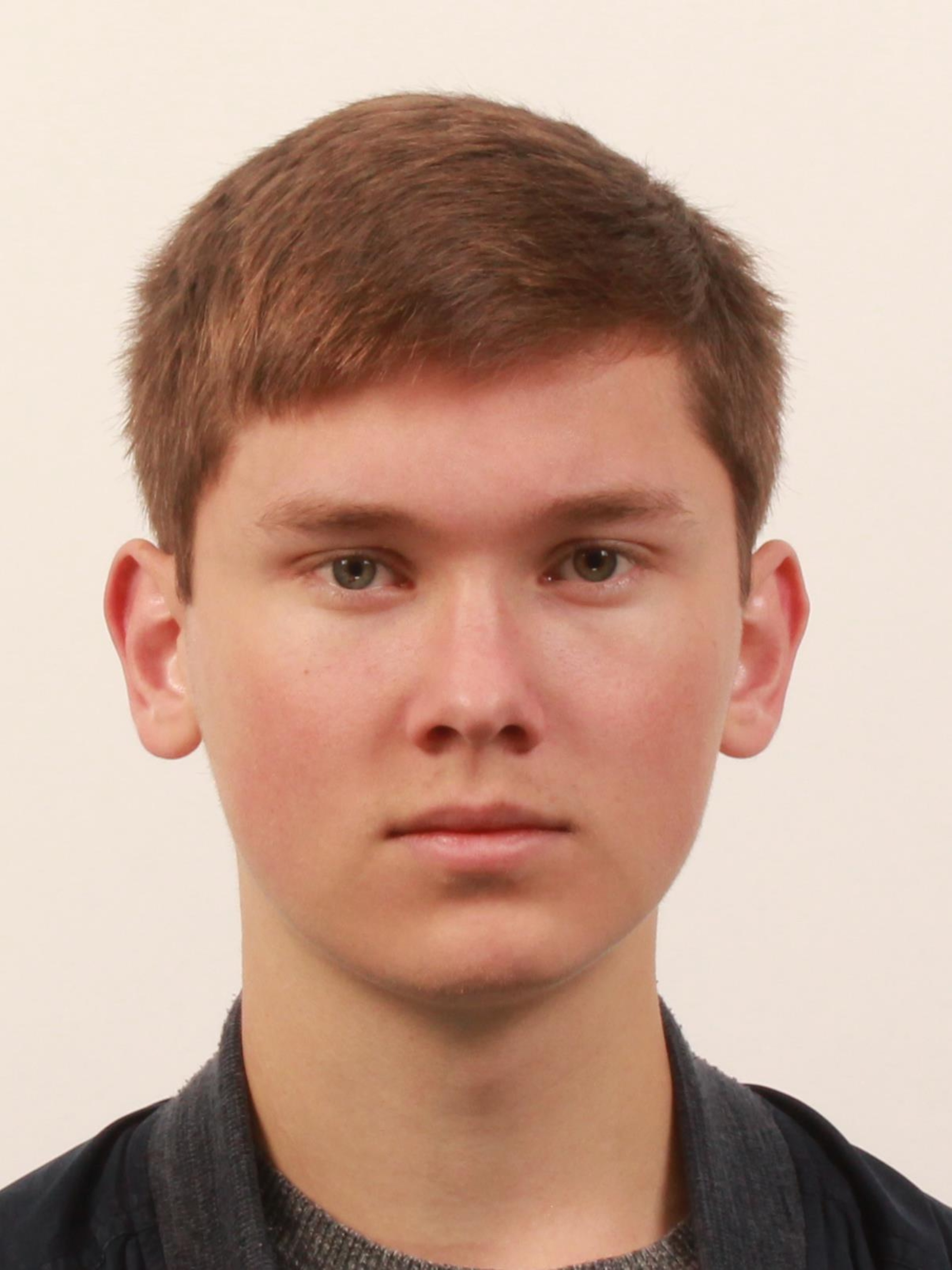}
    \label{fig:myfig_a}%
}
~
\subfigure[Morphed face image with alpha factor = 0.5]{
    \includegraphics[width=0.2\linewidth]{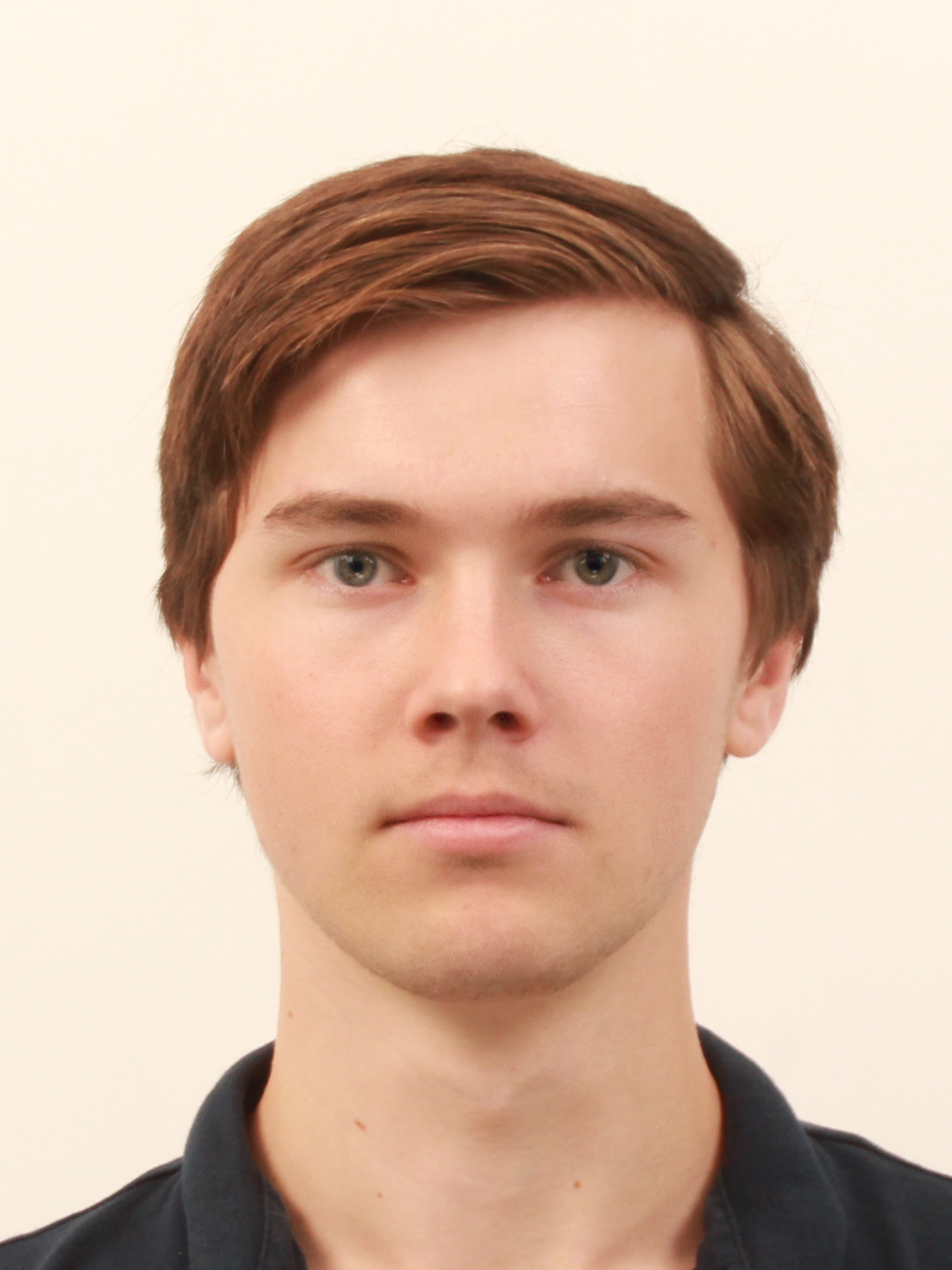}
    \label{fig:myfig_b}%
}
~
\subfigure[Morphed face image with alpha factor = 0.3]{
    \includegraphics[width=0.2\linewidth]{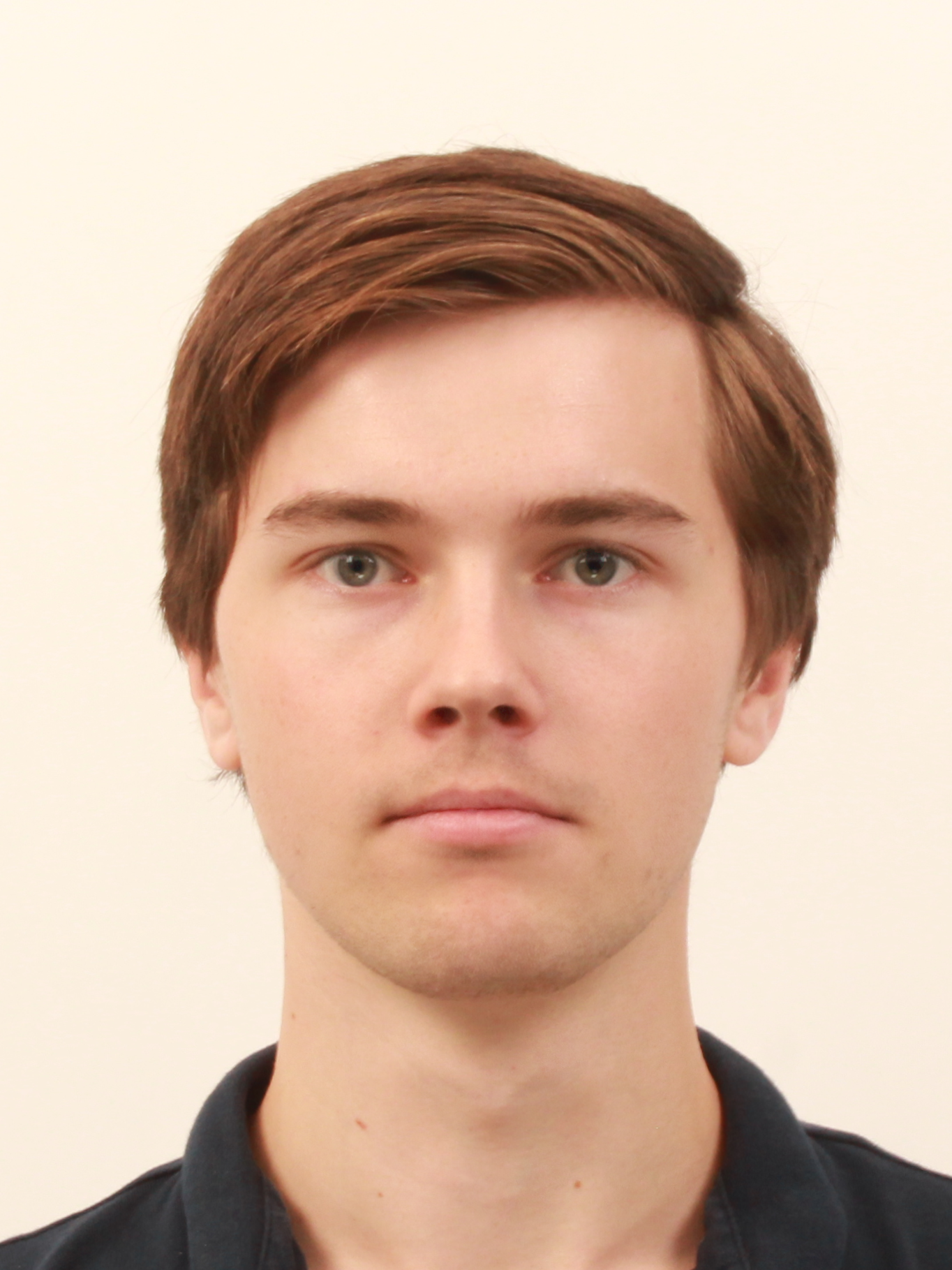}
    \label{fig:myfig_c}%
}
~
\subfigure[Bona fide face image (Contributor B)]{
    \includegraphics[width=0.2\linewidth]{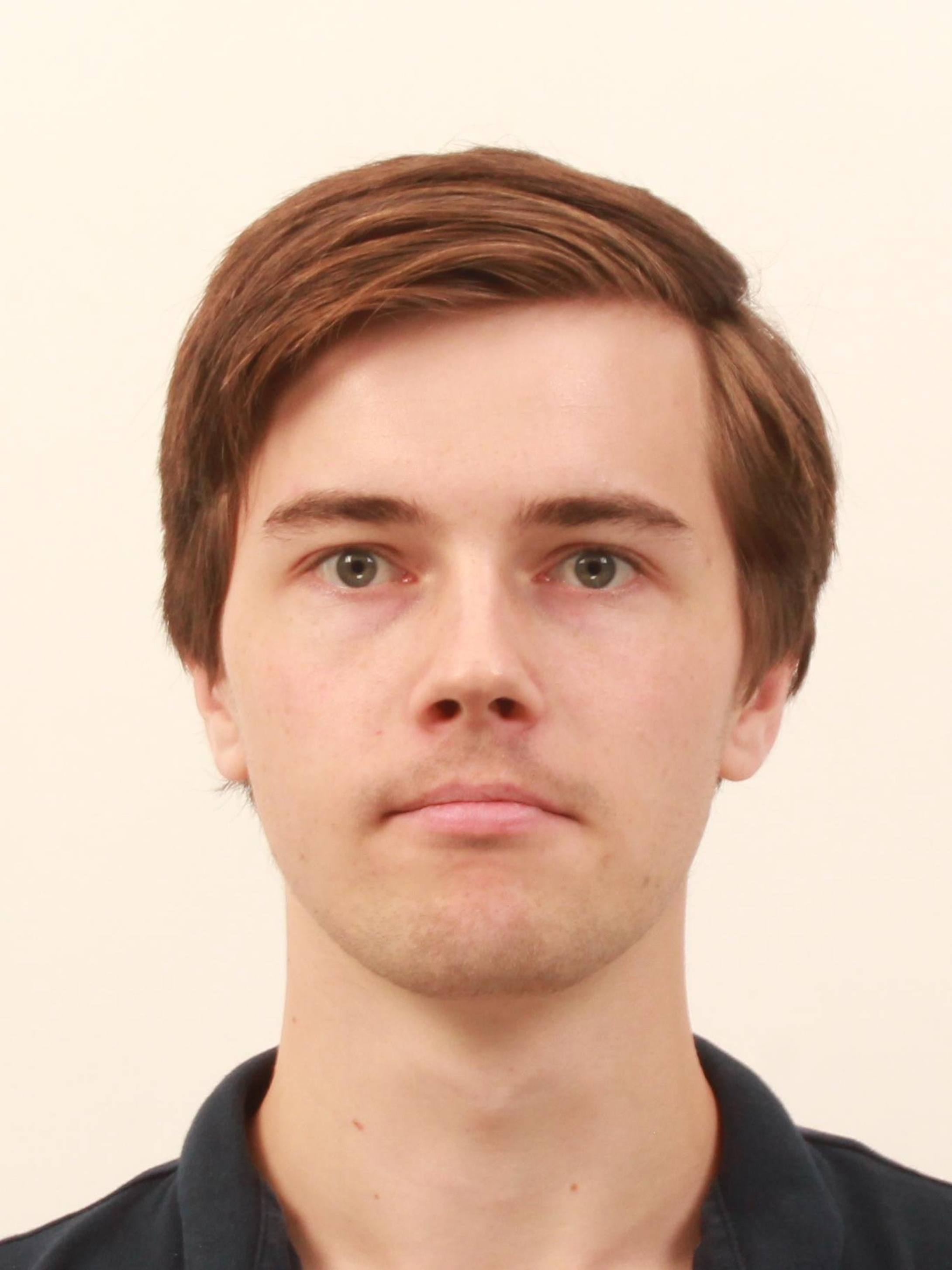}
    \label{fig:myfig_d}%
}
\vspace{-5mm}
\caption{Impact of morphing factors ($\alpha$) on morphing.}
  \label{fig:morphed-images}
\end{center}
\end{figure}

Unlike the noted previous works in Table \ref{tab:related-works}, the newly created morphed set in the SOTAMD database has a wide variation of employed morphing processes. Specifically, the morphing set consists of an unprocessed image set and fully-processed image set.
To increase the challenging nature of the dataset and in order to simulate realistic data, the post-processed images are printed and scanned using different pipelines. To further increase the diversity, each image pair was morphed using contributing factors (referred as alpha factor) of 0.3 and 0.5 for each of the two contributing faces. Examples of two morphed face images are shown in Figure ~\ref{fig:morphed-images}.

\begin{figure}
\begin{center}
\subcapcentertrue
\subfigure[Both automatically and manually post-processed digital morphed face image]{
    \includegraphics[width=0.4\linewidth]{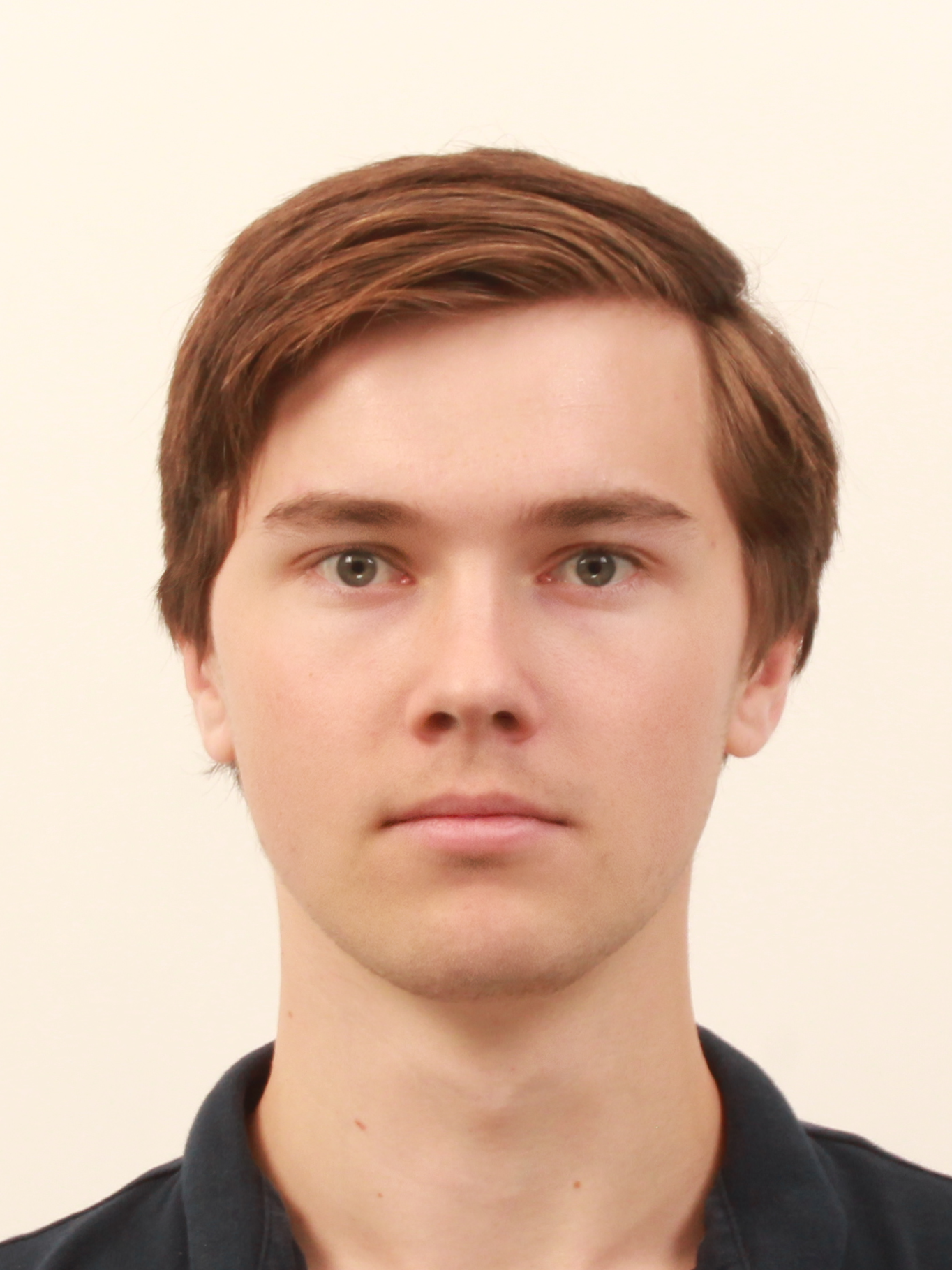}
    \label{fig:digital}%
}
~
\subfigure[Image ~\ref{fig:digital} but printed, scanned and compressed]{
    \includegraphics[width=0.4\linewidth]{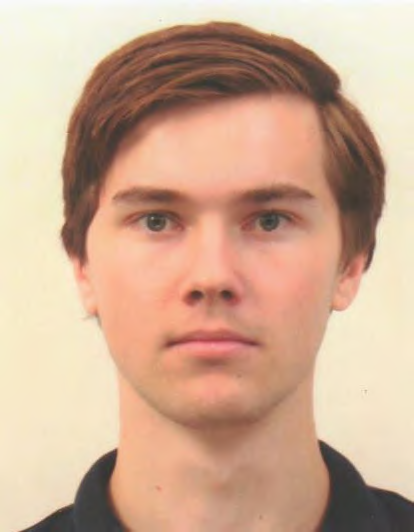}
    \label{fig:printed-scanned-compressed}%
}
\vspace{-5mm}
\caption{Illustration of post-processing - Careful processing to remove the artifacts can be noted in the eyelids, iris and nostril regions to eliminate the traces of the morphing process. Refer Figure~\ref{fig:Manual-Post-processing} for detailed illustration.}
 \label{fig:Post-processing}
\end{center}
\end{figure}

Furthermore, the processed images are resized using the OpenCV library \cite{opencv-slibrary} to maintain the same inter-eye distance distribution as observed in the morphed images to avoid any possibility of inferring the image class from it's dimensions. Post-processing methods consist of automatic and/or manual methods to conceal visible, and sometimes easy to detect morphing traces. Due to such variation in algorithms, any MAD algorithm that can achieve significant accuracy of detection on the SOTAMD dataset can be deemed as robust. Examples of automatically and manually post-processed digital morphed face image (left), and the same image after printing and scanning (right) are shown in Figure ~\ref{fig:Post-processing}. 

\begin{figure}
\begin{center}
\subcapcentertrue
\subfigure[Before]{
    \includegraphics[width=0.45\linewidth]{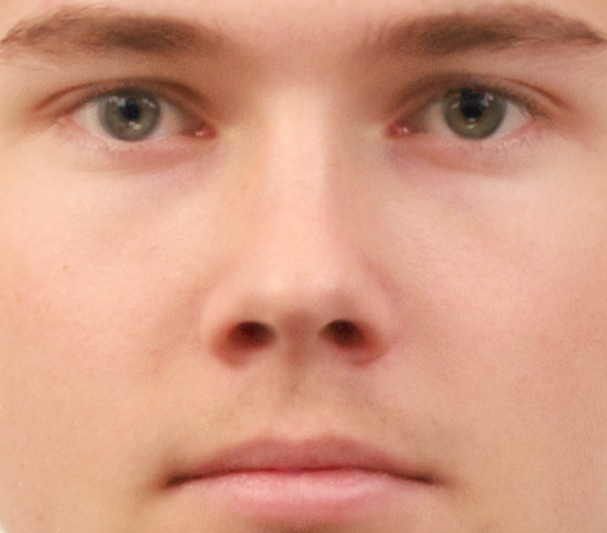}
    \label{fig:before}%
}
~
\subfigure[After]{
    \includegraphics[width=0.45\linewidth]{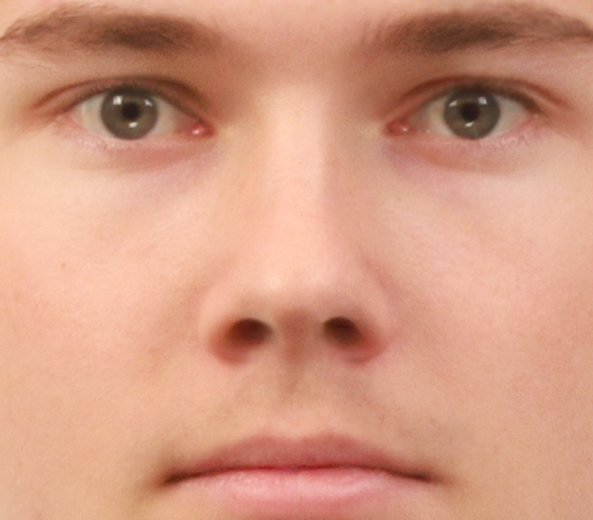}
    \label{fig:after}%
}
\vspace{-5mm}
\caption{Morphed face image before and after manual Post-processing from Figure~\ref{fig:Post-processing}. Only the central part of the face is reported to better appreciate the effect of artifact removal. Careful processing to remove the artifacts can be noted in the eyelids, iris and nostril regions to eliminate the traces of morphing process.}
 \label{fig:Manual-Post-processing}
\end{center}
\end{figure}

Examples of a morphed face image, before (left) and after (right) manual post-processing are shown in Figure ~\ref{fig:Manual-Post-processing}. Morphed face images that were both automatically and manually post-processed compose the most challenging subset. All the enrolment face images (bona fide and morphed) were processed with ICAO compliance \cite{ICAO-9303-p9-2015} testing software before entering into the database. An overview of the basic subsets of morphed face images is shown in Table ~\ref{tab:morphs}.\\
A detailed account of the morphing methods that were contributed by each partner can be seen in Table \ref{tab:morphing-dataset-methos-processing} which provides the various approaches used for automated and manual post-processing pipelines.

A subset of the generated morphed images has been printed and scanned using multiple pipelines (in analogy with the bona fide enrolment images); the number of morphed images in the database is therefore 2045 in digital format and 3703 printed and scanned.

\subsection{Gate images}
\begin{figure}[htbp]
\begin{center}
\subcapcentertrue
\subfigure[Mock ABC gate (UTW)]{
    \includegraphics[width=0.26\linewidth]{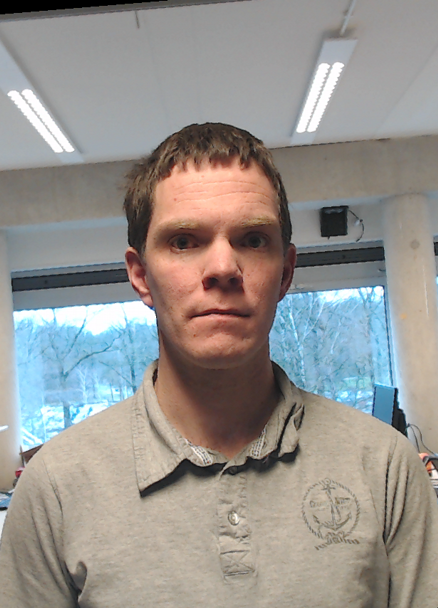}
}
~
\subfigure[ABC gate (HDA)]{
    \includegraphics[width=0.26\linewidth]{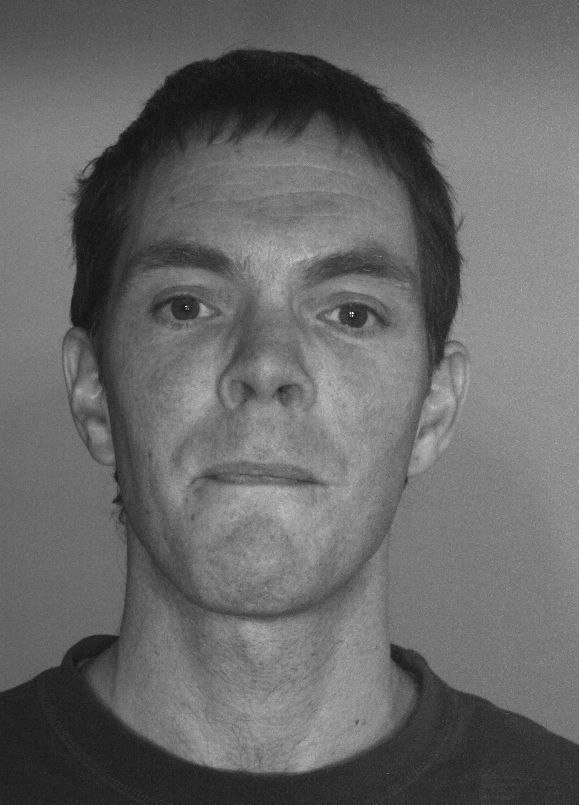}
}
~
\subfigure[IDEMIA’s MFace gate (NTN)]{
    \includegraphics[width=0.26\linewidth]{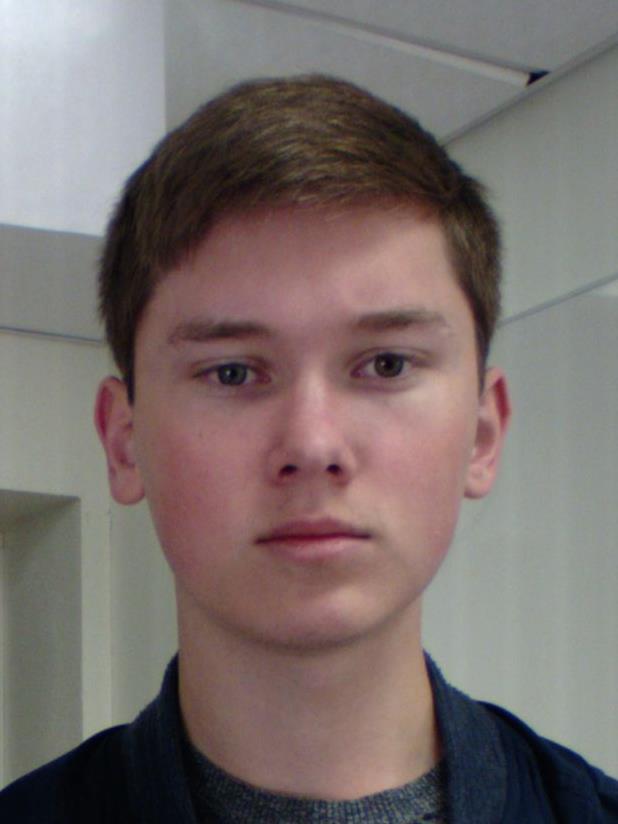}
}
\vspace{-5mm}
  \caption{Examples of probe face images captured from different ABC set-up.}
  \label{fig:probe-images}
\end{center}
\end{figure}

The SOTAMD database contains 10 gate images captured from each subject (overall 1500 images) during a single acquisition session at different locations under a simulated ABC gate operational scenario \footnote{Due to operational concerns not to interfere border control processes the images were not acquired with operational ABC gates at airport locations. Instead, HDA and NTN used a mock ABC gate setup provided by an ABC manufacturer, whereas UTW created a mock ABC gate setup.}.\\ 
As an additional measure, the quality of the images captured in the emulate ABC set-up was validated by reading the corresponding eMRTD chip images and verifying them against the captured gate image using COTS FRS.\\
The gate images were captured at two different partner facilities (Norwegian University of Science and Technology - referred to as NTN and Hochschule Darmstadt - referred to as HDA) from 100 subjects that directly corresponds to real ABC gates from two different vendors. These probe images that are generated from two different vendors capture devices, represent images that are used in real operational settings.
Another set (from University of Twente - referred to as UTW) of gate images from 50 subjects are captured with a simulated custom-built mock ABC gate. Thus, given three different set-ups of ABC gates, the probe-set provides a variation for benchmarking different MAD algorithms, which demands an agnostic nature and robustness of the algorithms. Examples of the different probe images captured from different set-up are illustrated in Figure ~\ref{fig:probe-images}.

\section{Evaluation Platform}
\label{sec-evaluation-platform}
We further present a new independent evaluation framework to measure the robustness of MAD. The MAD benchmarks have been realized following the testing framework of FVC-onGoing \cite{dorizzi2009FVConGoing, FVConGoing}. A web-based automated evaluation platform has been designed to track the advances in MAD, through continuously updated independent testing and reporting of performances on given benchmarks. FVC-onGoing benchmarks are grouped into benchmark areas according to the (sub)problem addressed and the evaluation protocol adopted (e.g. Fingerprint Verification, Palmprint Verification, Face Image ISO Compliance Verification, etc.). To maximize trustworthiness of the results, tests are carried out using a strongly supervised approach on a collection of sequestered datasets and results are reported on-line by using well known performance indicators and metrics. We follow the same design principles to evaluate the MAD algorithms in this work.

\begin{figure*}[!ht]
	\centering
	\includegraphics[width=0.75\textwidth]{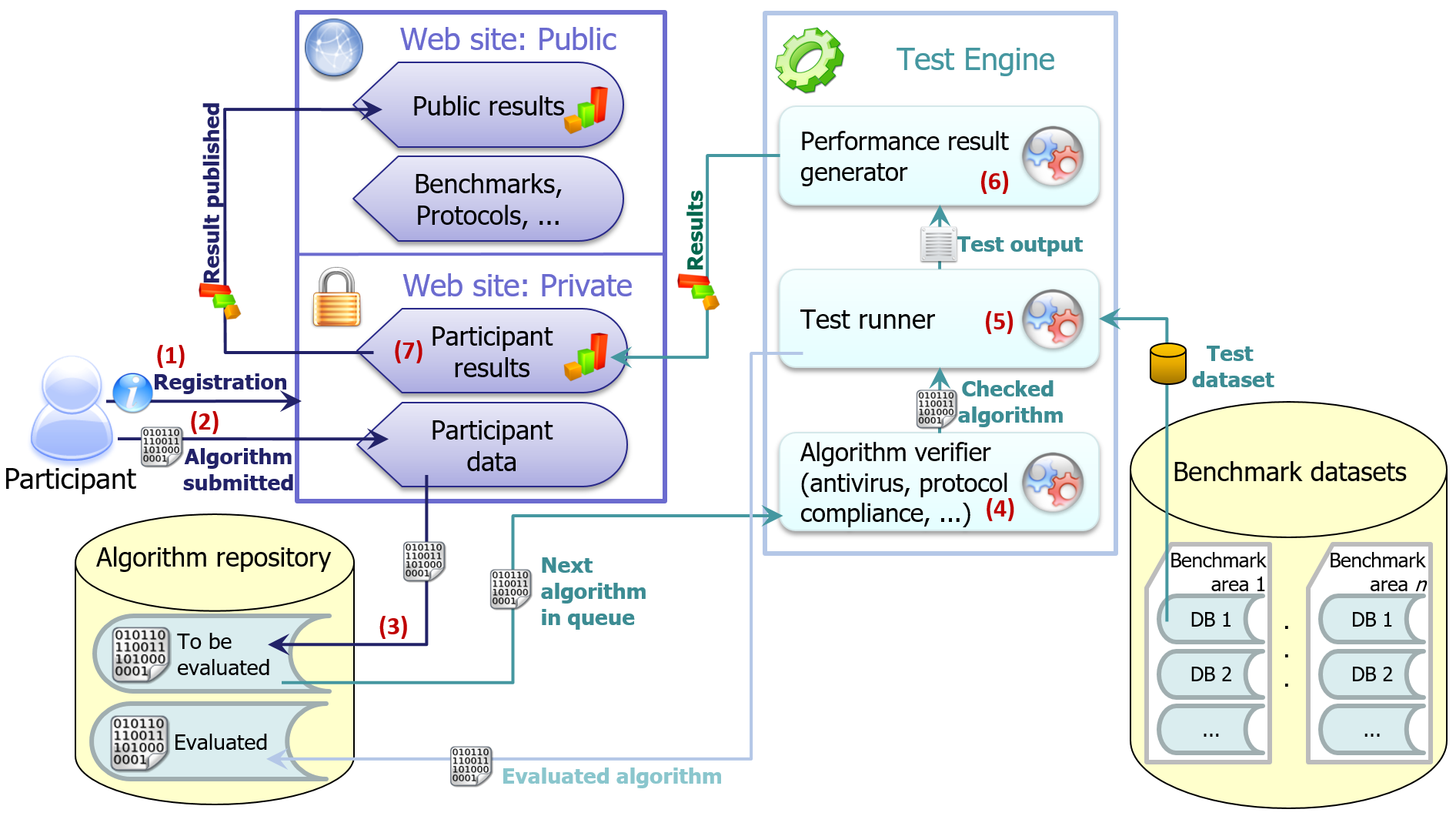}
	\hfill 
	\caption{The figure shows the architecture of the FVC-onGoing evaluation framework and an
example of a typical workflow: a given participant, after registering to the Web Site (1), submits
some algorithms (2) to one or more of the available benchmarks; the algorithms (binary
executable programs compliant to a given protocol) are stored in a specific repository (3). Each
algorithm is evaluated by the Test Engine that, after some preliminary checks (4), executes it on
the dataset of the corresponding benchmark (5) and processes its outputs (e.g. comparison scores)
to generate (6) all the results (e.g. EER, score graphs), which are finally published (7) on
the Web Site.}
	\label{fig:FVConGoingFramework}
\end{figure*}

The evaluation process is fully automated as illustrated in Figure \ref{fig:FVConGoingFramework} which consists of participant registration, algorithm submission, performance evaluation, and results visualization.
To protect sensitive information (biometric data) and to prevent external attacks, the FVC-onGoing framework is composed of two different modules physically located in two separate servers:
\begin{itemize}[leftmargin=*]
    \item The \textbf{Front-End server} containing the web site and the algorithm repository. 
    \item The \textbf{Test Engine server} containing the test engine and the benchmark datasets.
\end{itemize}

A firewall protects the Test Engine server by blocking all inbound and outbound connections on public and private networks.
Only a few authorized users can access the Test Engine server from a specific terminal using a protected local connection. Moreover, to avoid undesirable behaviour of the submitted algorithms, all of them are first analysed by antivirus software and then executed in a strongly controlled environment with minimal permissions.\\
Algorithms can be provided in the form of i) a Win32 console application or ii) a Linux dynamically-linked library compliant to NIST FRVT MORPH specifications \cite{NistFrvtMorph}.

\begin{table*}[b]
    \centering
    \caption{D-MAD and S-MAD benchmarks}
    \resizebox{0.9\textwidth}{!}{
        \begin{tabular}{|c|l|c|c|r|r|r|r|r|r|r|}
        \hline
        \rowcolor[rgb]{ .851,  .851,  .851} 
        \cellcolor[rgb]{ .851,  .851,  .851} & \multicolumn{1}{c|}{\cellcolor[rgb]{ .851,  .851,  .851}} & \cellcolor[rgb]{ .851,  .851,  .851} & \cellcolor[rgb]{ .851,  .851,  .851} & \multicolumn{5}{c|}{\cellcolor[rgb]{ .851,  .851,  .851}\textbf{Eye distance}} & \multicolumn{1}{c|}{\cellcolor[rgb]{ .851,  .851,  .851}} & \multicolumn{1}{l|}{\cellcolor[rgb]{ .851,  .851,  .851}} \\ \cline{5-9}
        \rowcolor[rgb]{ .851,  .851,  .851} 
        \multirow{-2}{*}{\cellcolor[rgb]{ .851,  .851,  .851}\textbf{\begin{tabular}[c]{@{}c@{}}Benchmark\\ area\end{tabular}}} & \multicolumn{1}{c|}{\multirow{-2}{*}{\cellcolor[rgb]{ .851,  .851,  .851}\textbf{Benchmark}}} & \multirow{-2}{*}{\cellcolor[rgb]{ .851,  .851,  .851}\textbf{Format}} & \multirow{-2}{*}{\cellcolor[rgb]{ .851,  .851,  .851}\textbf{\begin{tabular}[c]{@{}c@{}}Morphing\\ factor\end{tabular}}} & \multicolumn{1}{l|}{\cellcolor[rgb]{ .851,  .851,  .851}\textbf{Min}} & \multicolumn{1}{l|}{\cellcolor[rgb]{ .851,  .851,  .851}\textbf{Q\textsubscript{25}}} & \multicolumn{1}{l|}{\cellcolor[rgb]{ .851,  .851,  .851}\textbf{Q\textsubscript{50}}} & \multicolumn{1}{l|}{\cellcolor[rgb]{ .851,  .851,  .851}\textbf{Q\textsubscript{75}}} & \multicolumn{1}{l|}{\cellcolor[rgb]{ .851,  .851,  .851}\textbf{Max}} & \multicolumn{1}{c|}{\multirow{-2}{*}{\cellcolor[rgb]{ .851,  .851,  .851}\textbf{\begin{tabular}[c]{@{}c@{}}Bona fide\\ attempts\end{tabular}}}} & \multicolumn{1}{l|}{\multirow{-2}{*}{\cellcolor[rgb]{ .851,  .851,  .851}\textbf{\begin{tabular}[c]{@{}l@{}}Morph \\ attempts\end{tabular}}}} \\ \hline
         & D-MAD-SOTAMD\_D-1.0 & Digital & 0.3 and 0.5 & 80 & 156 & 311 & 515 & 1020 & 3000 & 30550 \\ \cline{2-11} 
        \multirow{-2}{*}{D-MAD} & D-MAD-SOTAMD\_P\&S-1.0 & Printed \& Scanned & 0.3 and 0.5 & 80 & 105 & 115 & 140 & 360 & 10960 & 55530 \\ \hline
        \multicolumn{11}{|l|}{} \\ \hline
         & S-MAD-SOTAMD\_D-1.0 & Digital & 0.3 and 0.5 & 90 & 326 & 456 & 533 & 1020 & 300 & 2045 \\ \cline{2-11} 
        \multirow{-2}{*}{S-MAD} & S-MAD-SOTAMD\_P\&S-1.0 & Printed \& Scanned & 0.3 and 0.5 & 80 & 105 & 111 & 138 & 170 & 1096 & 3703 \\ \hline
        \end{tabular}
    }
    \label{tab:S-MAD_D-MAD_benchmarks}%
\end{table*}

Two different benchmark areas (D-MAD and S-MAD) have been created to evaluate the accuracy of MAD algorithms in the differential- and single-image scenarios. Table \ref{tab:S-MAD_D-MAD_benchmarks} provides detailed information on the benchmarks contained in the two benchmark areas. Algorithms submitted to these benchmarks must comply to specific protocols, whose details are given on the FVC-onGoing web site \cite{FVConGoing}.

\subsection{Detection performance evaluation}
The evaluation platform is designed to report a number of performance metrics for MAD algorithms as detailed in this section. For each experiment bona fide and morphed face images are used to compute the Bona fide Presentation Classification Error Rate (BPCER) and the Attack Presentation Classification Error Rate (APCER). As defined in \cite{ISO_IEC_TestingAndReporting} the BPCER is the proportion of bona fide presentations falsely classified as morphing presentation attacks while the APCER is the proportion of morphing attack presentations falsely classified as bona fide presentations. The following performance indicators are reported:
\begin{itemize}[leftmargin=*]
\item EER (detection Equal-Error-Rate): the error rate for which BPCER and APCER are identical 
\item BPCER\textsubscript{10}: the lowest BPCER for APCER$\leq$10\%
\item BPCER\textsubscript{20}: the lowest BPCER for APCER$\leq$5\%
\item BPCER\textsubscript{100}: the lowest BPCER for APCER$\leq$1\%
\item REJ\textsubscript{NBFRA}: Number of bona fide face images that cannot be processed
\item REJ\textsubscript{NMRA}: Number of morphed face images that cannot be processed
\item Bona fide and Morph detection score distributions
\item APCER(\textit{t})/BPCER(\textit{t}) curves, where \textit{t} is the detection threshold
\item DET(\textit{t}) curve (the plot of BPCER against APCER)
\end{itemize}

\subsection{Protocols for Evaluation}
\label{sec:evaluation-protocols}
In order to benchmark the MAD algorithms, we defined two specific protocols for D-MAD and S-MAD respectively:
\begin{itemize}[leftmargin=*]
    \item \textit{D-MAD}: in this case, the algorithms receive as input a pair of images (an enrolment image and a gate image) and are requested to estimate the probability that the enrolment image is morphed, based on a differential analysis of the two input images. The enrolment images available in the database are thus compared against the gate images (i.e. trusted live capture) according to the following protocol:
    \begin{itemize}[leftmargin=*]
        \item \textbf{Bona fide images}: the bona fide enrolment image is compared against the gate images of the same subject;
        \item \textbf{Morphed images (factor 0.3)}: the morphed enrolment image is compared against the gate images of the subject who contributed least in the morphing (the hidden identity);
        \item \textbf{Morphed images (factor 0.5)}: the morphed enrolment image is compared against the gate images of both contributing subjects.
    \end{itemize}
        \item \textit{S-MAD}: in this case, the algorithms receive as input a single image and are requested to estimate the probability that the image is morphed (i.e. to report a morphing likelihood score). To this aim, the probe set consists of the whole set of available enrolment images (bona fide and morphed). 
\end{itemize}
The resulting number of attempts for the two benchmarks is provided in Table \ref{tab:S-MAD_D-MAD_benchmarks}.
\section{MAD Algorithms}
\label{sec:mad-algoritms}
A number of existing state of the art MAD algorithms are evaluated on the newly created SOTAMD database using the new evaluation platform. Within the scope of this work, both D-MAD and S-MAD algorithms have been submitted to the corresponding FVC-onGoing benchmarks.  In this section, we provide a brief description of the algorithms that were tested on the newly developed database and the evaluation platform.

\subsection{D-MAD}
A D-MAD algorithm uses additional information from a second image known to be bona fide (e.g. a live image captured in an ABC gate) to detect morphed face images. D-MAD algorithms obtain the differences in images using textural features (textural features or deep features) or landmark shifts. We present a set of D-MAD algorithms evaluated on \Rev{SOTAMD} database in the subsequent sections.

\subsubsection{BSIF}
It is based on a set of texture features obtained using the Binarized Statistical Independent Features (BSIF) with a 8-bit filter of size $3x3$, applied on the normalized and aligned image \cite{scherhag2018towards}. Given the histogram feature vector of the dimension $1\times4096$ for  $h_s$ and $h_t$ respectively, the difference is presented to a \Rev{pre-trained} SVM classifier trained on the bona fide and morphed data from FERET \cite{phillips1998feret} and FRGC \cite{phillips2005overview} images. The approach also considers a number of post-processing steps such as median filtering, histogram normalization and sharpness processing on the images before training the SVM classifier for morphs generated from FaceMorpher and OpenCV. 

\subsubsection{DFR}
It utilizes the information of the embeddings (feature vectors) of the ArcFace algorithm \cite{Deng2018}, a ResNet based face recognition system. The fundamental idea is to use the feature vectors of the face-generating neural network to train an SVM. Since the neural network does not encounter morphed facial images during training, it can be excluded that the feature extraction overfits to artifacts of certain morphing algorithms, which in turn leads to a higher robustness of the resulting MAD algorithm. The ArcFace feature vector has a length of 512 features. The feature vectors of the e-gate live capture and the suspected morph image are subtracted. The resulting difference is used to train an SVM with RBF kernel. The algorithm evaluated in this paper was trained on the bona fide and morphed data from FERET \cite{phillips1998feret} and FRGC \cite{phillips2005overview}. Details of the DFR MAD algorithm can be found in \cite{scherhag2020deep}.

\subsubsection{MBLBP}
It consists of pre-processing, calculation of multiple block LBP from both $I_s$ and $I_t$ followed by classifying them as a bona fide image or morphed image using the pre-trained SVM classifer \cite{scherhag2018towards}. The Dlib landmark detector is used to detect the facial area and the  landmarks with the face in the pre-processing step where the face is realigned and normalized to achieve ICAO compliance \cite{ICAO-9303-p9-2015}. The normalised face image is then cropped to the $320\times320$ pixel wide region of  from which the LBP information is extracted using $4\times4$ equally sized blocks of the image. Within each block, a window size of $5\times5$ pixels is employed to obtain the histograms. Given the histogram of $h_s$ and $h_t$ for $I_s$ and $I_t$ respectively, a difference of $h_s$ and $h_t$ is obtained which is given to the SVM classifier to obtain a final decision on suspected image as morphed or bona fide image. Details on the MBLBP algorithm can be found in \cite{scherhag2018towards}.

\subsubsection{WL}

This method is based the fact that facial landmarks are usually averaged between two individuals when morphed images are created. Therefore, the distance of a given landmark (e.g., right corner of the right eye) between two bona fide images of the same subject will be smaller than the distance between that same landmark from a bona fide images of the subject and the morphed images with another subject. To exploit this idea, a set of 68 facial landmarks is extracted from each input image using dlib. Subsequently, two types of features are computed: Euclidean distances between landmarks, and angles between a pre-defined set of neighbouring landmarks. In order to account for the reliability of the landmarks estimation (e.g., the eye corners are more stable than landmarks on the lips), different weights are applied to the distances before they are classified as bona fide or morphed images using an SVM. Details on the computations of the distances and angles can be found in \cite{scherhag2017landmarks,damer2018detecting}.

\subsubsection{DR}
This method is based on the differentiating the image from bona fide image captured from trusted environment, (e.g., ABC gate) and the suspected image from Machine-Readable Travel Document (eMRTD) \cite{mohan2019robust}. Both images $I_s$ and $I_t$ are decomposed into the normal maps, and diffuse map using SfSNet \cite{sengupta2018sfsnet} following which the diffuse reconstructed image and a quantized normal map are obtained. From the diffuse map, the features are extracted using ‘fc7’ activation layer of AlexNet \cite{krizhevsky2012imagenet}. The features from the normal map are extracted by converting them to quantized spherical angles (quantization is 24-bit). The features are used to train polynomial SVM classifiers  for each set of features. The classifiers are used then used to determine if the suspected image is morphed or not based on the fusion of scores from each individual classifier corresponding to normal map and diffuse map. Details on the DR D-MAD algorithm can be found in \cite{mohan2019robust}.

\subsubsection{Face demorphing}
The idea of Face Demorphing (FaDe) \cite{Ferrara2018demorphing} involves inverting the morphing process in a reverse engineered manner.  Given a suspected image $I_s$ that is corresponding to image stored in the ID document where $I_s$ is generally a linear combination of multiple images. $I_m=I_a+I_c$ where $I_a$ and $I_c$ are the face images of bona fide accomplice and a criminal respectively. The assumption on the other end is that for a genuine ID document (with no morphing attack) the image $I_m$ is a combination of two identical images (for e.g., $I_m=I_a+I_a$), where $I_a$ \Rev{is the bona fide image}. 

Given the captured image $I_t$ in a trusted environment, demorphing algorithm obtains a difference between the suspected image $I_s$ and the captured image $I_t$ to obtain a demorphed image $I_d$. When the $I_d$ is compared against the $I_t$ using a FRS system, a high comparison score ($S$) indicates no morphing and lower score indicates higher probability of morphing. Ferrara et al. \cite{Ferrara2018demorphing} employ Dlib for comparing the trusted capture image $I_t$ and demorphed image $I_d$ as given below:
\begin{equation}
\centering
 S =\begin{cases}
    \max\biggl[0, \dfrac{(d-\tau_1)}{(2\times(\tau_2 - \tau_1))}  \biggr], & \ if \ d\leq \tau_2 \\
    \max\biggl[1, 0.5 + \dfrac{(d-\tau_2)}{(2\times(\tau_3 - \tau_2))}  \biggr], & \text{otherwise}.
  \end{cases}
  \label{eqn:demorphing}
\end{equation}

where $\tau_1, \tau_2, \tau_3$ are thresholds chosen om empirical trials set to $0.3699, 0.4565, 0.5469$ respectively.

\subsection{S-MAD}
An S-MAD algorithm determines whether an image is morphed directly i.e. without using a trusted reference image. Most of the S-MAD algorithms first extract the features from the suspected image using textural or deep networks, followed by learning a classifier. The learnt classifier is used to determine if the image is morphed or not. We briefly describe the set of S-MAD algorithms evaluated in this work.

\subsubsection{PRNU}
This algorithm is based on the analysis of Photo Response Non-Uniformity (PRNU). In essence, the PRNU stems from slight variations among individual pixels during the photoelectric conversion in digital image sensors. As a consequence, it is present in all acquired images and can be considered as an inherent part of any sensor's output. In fact, the PRNU has been successfully used for different forensic tasks, such as device identification or detection of digital forgeries. For the particular purpose of detecting morphed images \cite{scherhag2019detection}, the PRNU is extracted from the preprocessed facial images and subsequently split into cells. From each cell, the variance of 100-bin histograms of the PRNU is computed. Then, the minimum value among all cells is thresholded to obtain a bona fide vs.~morphed image decision. More details on this MAD mechanism can be found in \cite{scherhag2019detection}.

\subsubsection{Scale-Space Ensemble Approach (SSE)}
The algorithm is based on ensemble approach of extracting textural features followed by learning a classifier\cite{raghavendra2018detecting}. With the set of scores obtained from different classifiers learnt from different features, the final decision is made on whether the image is bona fide or morphed. Specifically, the image is decomposed in different color spaces such as YCbCr and HSV space. For each channel of the color space, the image is decomposed into different scale spaces using a Laplacian pyramid with 3 level decomposition. Further different textural features using Binarized Statistical Independent Features (BSIF), Local Binary Patterns (LBP) and Histogram of Gradients (HOG) are obtained. The obtained features are further used to learn the Collaborative Representative Classifier (CRC). While the testing is carried out on the SOTAMD dataset, the training was performed on a dataset derived from the FRGC face dataset. More details can be found in \cite{raghavendra2017transferable}.

\subsubsection{Deep-S-MAD}
This algorithm uses well-known pre-trained CNNs to detect morphed images \cite{ferrara2019face}. Pre-trained networks have been fine tuned using a large set of artificially generated digital images (both bona fide and morphed). Moreover, in order to deal with the print and scan process (P\&S), a further fine tuning step has been performed for the P\&S case exploiting a set of images artificially generated to simulate P\&S. The simulation follows a mathematical model that allows to control different image characteristics, related to both image visual quality and low-level signal content. In particular, the main visual effects produced when an image is printed and scanned can be successfully reproduced (blurring, gamma correction, color adjustment or noise).\\ The AlexNet architecture pre-trained on ImageNet \cite{krizhevsky2012imagenet} has been used on digital images while the VGG-Face16 \cite{parkhi2015deep} architecture pre-trained on the VGG-Face dataset \cite{parkhi2015deep} has been used on P\&S images.

\subsubsection{S-MBLBP}
\label{sssec:s-mblbp}
{The created classification system extracts multi-block local binary patterns from a face image and uses a support vector machine with a linear kernel to classify it as either morphed or bona fide \Rev{\cite{scherhag2018towards}}. The approach optimises the feature extraction process by using uniform LBPs with radius, r = 1 (i.e. number of neighbours, n = 8), and a histogram layout of \Rev{$3\times3$}. Before feature extraction the face is detected and cropped with a HOG-based face detector \cite{dlib}, converted to grey scale and finally histogram equalization is applied to enhance image contrast. The \Rev{$3\times3$} histogram layout is realized by splitting the face image by 2 equidistant vertical and horizontal lines. A single histogram contains 59 feature values, which means that after concatenating the 9 histograms of our layout our feature space has 531 dimensions. The classifier was trained on \cite{phillips2005overview} and \cite{ist-eurecom}. As pre-processing steps, all training images were converted to png format without any compression to avoid jpg compression artefacts being detected, and resized using nearest neighbour interpolation to the average size of the three training datasets. Additionally, faces were horizontally aligned to make them similar to (ICAO compliant) benchmark images.}

\section{Results and Discussion}
\label{sec:results-discussion}
\subsection{Results -D-MAD}
The results observed in the \textit{Digital Image Benchmark (D-MAD-SOTAMD\_D-1.0)} are reported in Figure \ref{fig:DETPlotOverall_D} (also Table \ref{tab:overallRes_D} in Appendix for the results on two subsets with morphing factor 0.3 and 0.5 respectively). In particular, the DET plots in Figure \ref{fig:DETPlotOverall_D} refer to the overall results, additional results are reported in Appendix A.\\
\begin{figure}[!ht]
	\centering
	\includegraphics[width=0.85\textwidth]{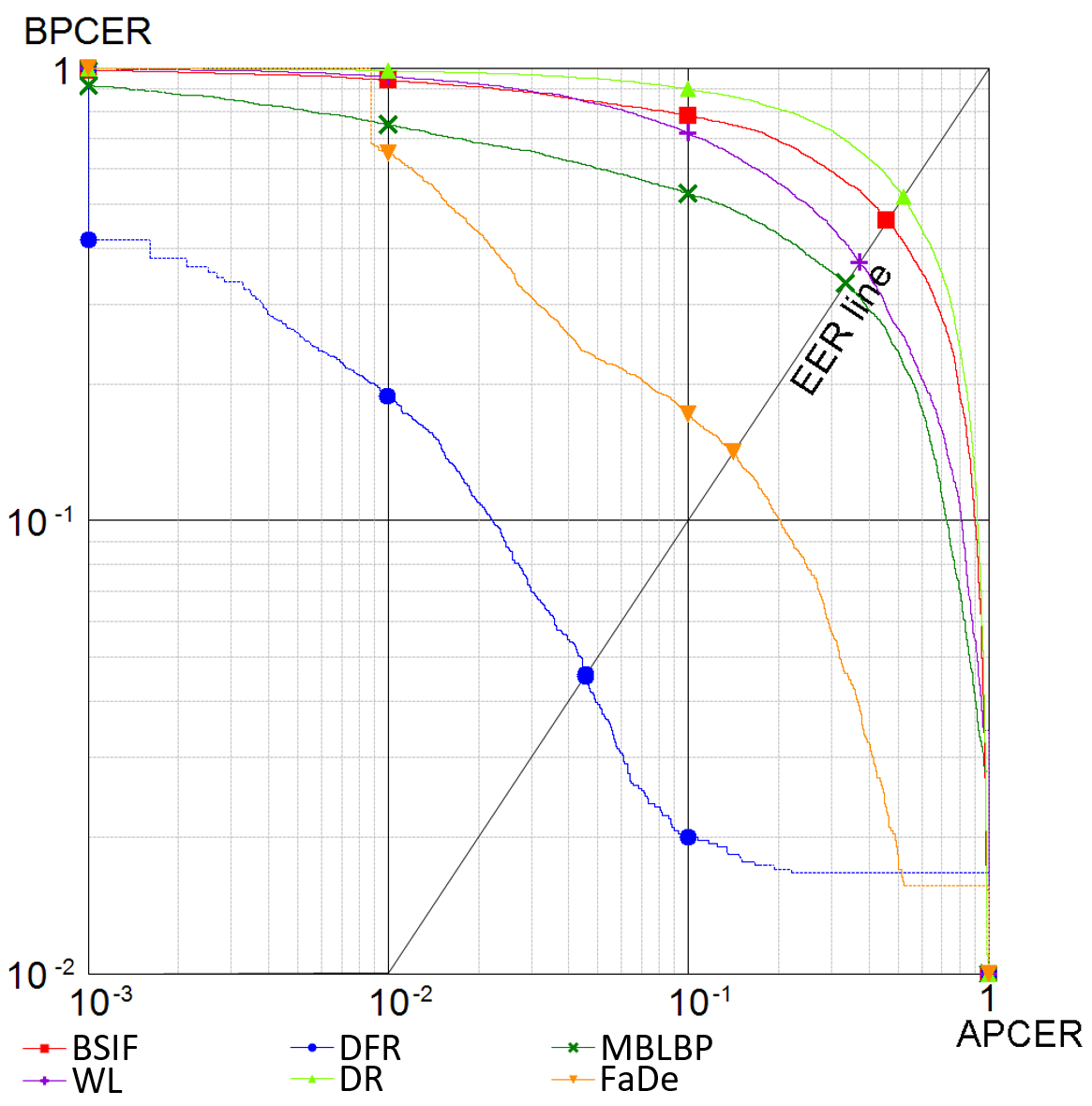}
	\hfill 
	\caption{DET plots for the D-MAD-SOTAMD\_D-1.0.}
	\label{fig:DETPlotOverall_D}
\end{figure}

The detection accuracy of some of the evaluated algorithms is quite modest. Two algorithms perform better than the average, and the algorithm DFR in particular reaches very promising results. The reason for the general under-performance of MAD algorithms with respect to the detection accuracy reported in the original publications could be due to the difficulty of the benchmark dataset and the over-specialization of said algorithms on the native training sets used previously in the research labs. As to the FaDe approach, its better generalization capability is probably due to the absence in the method of a specific training stage and/or hyperparameters tuning. The good performance of DFR can be attributed to the fact that the ArcFace algorithm used for feature extraction was trained independently of morphed images and thus the extracted feature vectors are not overfitted to the artifacts of individual MAD algorithms.\\
Table \ref{tab:overallRes_D} reports the performance of the tested MADs on the entire set of images as well as separately for the subsets of images with morphing factor 0.3 and 0.5. The results related to the morphing factor 0.3 are in general slightly better than those obtained on the entire database. A noticeable improvement can only be observed on all the performance indicators for DFR and FaDe algorithms. The behavior of FaDe is explainable if we consider that the algorithm has been designed to work on asymmetric morphings. The performance gain of the DFR can be attributed to the use of the difference vector. If the morphing factor is lower, the difference increases and so does the possibility to detect the morph.

For a deeper comprehension of the main image characteristics affecting to a larger extent the MAD performance, the results have been analyzed for specific subsets of images, described in Table \ref{tab:test_subsets} presented in Appendix. The subsets have been selected according to the number of images available (too small subsets are therefore discarded).

The degree of influence of each specific subset with respect to the overall performance has been evaluated computing, for each subset s, the percentage deviation between the EER measured on the specific subset ($eer_s$) and the EER measured on the whole set of images:
\begin{equation}
    dev_s=\frac{eer_s-eer_o}{eer_o}\times100
\end{equation}

A negative deviation indicates that the specific subset is “easier” with respect to the overall set of images (a lower EER value has been observed), high positive values identify more difficult subsets. The deviation computed for each algorithm, as well as the average deviation ($\overline{dev_s}$) for the subset of tests with morphing factor 0.3 are reported in Table \ref{tab:subsetRes_D0.3} in Appendix where the results are sorted by $\overline{dev_s}$. Some interesting results can be observed, in relation to the main attributes characterizing the database images:
\begin{itemize}[leftmargin=*]
\item \textit{Ethnicity}: in general the morphed images produced with Indian-Asian and Middle Eastern subjects are easier to detect for most of the algorithms. The cardinality of these subsets is lower than European/American, and the chance of selecting lookalike subjects for morphing was lower.
\item \textit{Automatic or manual post-processing}: as expected manual post-processing (i.e., retouching for artefact removal) makes morphing detection more difficult w.r.t. automatic post-processing, even if the difference is just minor here. 
\item \textit{Manual post-processing technique}: significant differences can be observed in relation to the manual post-processing executor, thus confirming the importance of manual retouching aimed at removing small artefacts; while PM03 and PM06 are easier to detect, especially for some algorithms, PM02 and PM05 are more difficult to spot. 
\item \textit{Subset of Morphs}: the subset containing UTW images is more difficult with respect to those from the other partners. In fact, in this case, very similar pairs of subjects were selected, making the resulting morphs more difficult to be detected.
\item \textit{Morph quality}: as expected high quality morphs (i.e., those accepted by commercial  face verification algorithms) are more difficult to detect than low quality morphs (i.e., those already rejected by face verification algorithms). 
\item \textit{Morphing algorithm}: the results over different morphing algorithms are quite different; algorithms C06, C07 and C03 are generally easier to detect, while C02 and C01 are quite hard for most of the D-MAD algorithms.
\item \textit{Age}: the results on subjects in the range 56-75 are generally much worse than those related to younger subjects; as per the Traits subsets (see below) we argue that the transfer of evident skin characteristics such as wrinkles, freckles or moles, can make the morphed images similar enough to both subjects. 
\item \textit{Gender}: morphing detection in female subjects looks on average more difficult.
\item \textit{Traits}: the error rate on images with specific traits (moles, freckles) is on average higher than that measured on images without particular facial traits. See the above discussion on Age. 
\end{itemize}
    \Rev{The results reported in Table \ref{tab:subsetRes_D0.3} (Appendix) show that, even if a common behaviour can be observed for several subsets, in a number of cases (e.g. Type of Post-processing or Ethnicity) different algorithms provide significantly different performance. This leads us to suppose that the tested D-MADs produce quite independent errors and a combination of such different techniques can lead to a performance improvement.}

\begin{table*}[htbp]
  \centering
  \caption{Performance indicators measured on the D-MAD-SOTAMD\_P\&S-1.0 benchmark for the overall set of images and for the subsets of images with morphing factor 0.3 and 0.5.}
  \resizebox{0.9\textwidth}{!}{
    \begin{tabular}{|c|c|c|p{6.22em}|p{4.055em}|p{4.055em}|p{4.055em}|p{4.78em}|p{4.055em}|p{4.055em}|}
    \hline
    \rowcolor[rgb]{ .851,  .851,  .851} \multicolumn{1}{|p{4.055em}|}{\textbf{Test}} & \multicolumn{1}{p{6.055em}|}{\textbf{Bona fide comparisons}} & \multicolumn{1}{p{5.945em}|}{\textbf{Morphed comparisons}} & \textbf{Algorithm} & \textbf{EER} & \textbf{BPCER\textsubscript{10}} & \textbf{BPCER\textsubscript{20}} & \textbf{BPCER\textsubscript{100}} & \textbf{REJ\textsubscript{NBFRA}} & \textbf{REJ\textsubscript{NMRA}} \\
    \hline
    \multirow{5}[10]{*}{Overall} & \multirow{5}[10]{*}{10960} & \multirow{5}[10]{*}{55530} & BSIF & 51.36\% & 95.66\% & 98.38\% & 99.55\% & 1.35\% & 1.92\% \\
\cline{4-10}          &       &       & DFR & \textbf{4.62\%} & \textbf{1.77\%} & \textbf{4.08\%} & \textbf{19.70\%} & 1.46\% & 2.11\% \\
\cline{4-10}          &       &       & MBLBP & 29.28\% & 51.50\% & 62.38\% & 81.16\% & 2.66\% & 3.56\% \\
\cline{4-10}          &       &       & WL & 36.17\% & 70.37\% & 82.75\% & 95.58\% & 3.47\% & 4.19\% \\
\cline{4-10}          &       &       & DR  & 50.13\% & 90.26\% & 95.37\% & 99.18\% & \textbf{0.00\%} & \textbf{0.00\%} \\
\cline{4-10}          &       &       & FaDe   & 17.22\% & 24.82\% & 32.37\% & 74.61\% & 0.16\% & 0.25\% \\
    \hline
    \multicolumn{10}{|c|}{} \\
    \hline
    \multirow{5}[10]{*}{0.3} & \multirow{5}[10]{*}{10960} & \multirow{5}[10]{*}{18530} & BSIF & 50.98\% & 95.60\% & 98.39\% & 99.56\% & 1.35\% & 1.93\% \\
\cline{4-10}          &       &       & DFR & \textbf{2.09\%} & \textbf{1.55\%} & \textbf{1.55\%} & \textbf{12.39\%} & 1.46\% & 2.13\% \\
\cline{4-10}          &       &       & MBLBP & 27.58\% & 47.03\% & 57.72\% & 75.76\% & 2.66\% & 3.63\% \\
\cline{4-10}          &       &       & WL & 31.83\% & 62.40\% & 76.43\% & 93.49\% & 3.47\% & 4.26\% \\
\cline{4-10}          &       &       & DR  & 50.38\% & 90.42\% & 95.64\% & 99.25\% & \textbf{0.00\%} & \textbf{0.00\%} \\
\cline{4-10}          &       &       & FaDe   & 11.25\% & 12.74\% & 20.56\% & 38.38\% & 0.16\% & 0.23\% \\
    \hline
    \multicolumn{10}{|c|}{} \\
    \hline
    \multirow{5}[10]{*}{0.5} & \multirow{5}[10]{*}{10960} & \multirow{5}[10]{*}{37000} & BSIF & 51.54\% & 95.67\% & 98.35\% & 99.55\% & 1.35\% & 1.92\% \\
\cline{4-10}          &       &       & DFR & \textbf{5.34\%} & \textbf{2.21\%} & \textbf{5.60\%} & \textbf{23.16\%} & 1.46\% & 2.09\% \\
\cline{4-10}          &       &       & MBLBP & 30.11\% & 53.28\% & 64.63\% & 83.56\% & 2.66\% & 3.53\% \\
\cline{4-10}          &       &       & WL & 38.15\% & 73.32\% & 84.90\% & 96.51\% & 3.47\% & 4.15\% \\
\cline{4-10}          &       &       & DR  & 49.96\% & 90.20\% & 95.22\% & 99.08\% & \textbf{0.00\%} & \textbf{0.00\%} \\
\cline{4-10}          &       &       & FaDe   & 19.68\% & 28.55\% & 38.46\% & 100.00\% & 0.16\% & 0.27\% \\
    \hline
    \end{tabular}%
    }
  \label{tab:overallRes_PS}%
\end{table*}%

\begin{table*}[htbp]
  \centering
  \caption{Performance indicators measured on the S-MAD-SOTAMD\_P\&S-1.0 benchmark for the overall set of images and for the subsets of images with morphing factor 0.3 and 0.5.}
  \resizebox{0.9\textwidth}{!}{
    \begin{tabular}{|c|c|c|l|r|r|r|r|r|r|}
    \hline
    \rowcolor[rgb]{ .851.  .851.  .851} \multicolumn{1}{|p{4.055em}|}{\textbf{Test}} & \multicolumn{1}{p{6.055em}|}{\textbf{Bona fide comparisons}} & \multicolumn{1}{p{5.945em}|}{\textbf{Morphed comparisons}} & \textbf{Algorithm} & \textbf{EER} & \textbf{BPCER\textsubscript{10}} & \textbf{BPCER\textsubscript{20}} & \textbf{BPCER\textsubscript{100}} & \textbf{REJ\textsubscript{NBFRA}} & \textbf{REJ\textsubscript{NMRA}} \\
    \hline
    \multirow{4}[8]{*}{Overall} & \multirow{4}[8]{*}{1096} & \multirow{4}[8]{*}{3703} & PRNU   & 48.04\% & \textbf{85.86\%} & \textbf{97.35\%} & 100.00\% & 0.09\% & 0.00\% \\
\cline{4-10}          &       &       & SSE  & 54.37\% & 94.89\% & 98.27\% & \textbf{99.91\%} & 0.00\% & 0.00\% \\
\cline{4-10}          &       &       & {Deep-S-MAD}   & \textbf{37.10\%} & 100.00\% & 100.00\% & 100.00\% & 0.00\% & 0.00\% \\
\cline{4-10}          &       &       & S-MBLBP   & 43.34\% & 100.00\% & 100.00\% & 100.00\% & 0.09\% & 0.00\% \\
    \hline
    \multicolumn{10}{|c|}{} \\
    \hline
    \multirow{4}[8]{*}{0.3} & \multirow{4}[8]{*}{1096} & \multirow{4}[8]{*}{1853} & PRNU   & 48.49\% & \textbf{86.13\%} & \textbf{97.17\%} & 100.00\% & 0.09\% & 0.00\% \\
\cline{4-10}          &       &       & SSE  & 55.18\% & 94.89\% & 98.36\% & \textbf{99.91\%} & 0.00\% & 0.00\% \\
\cline{4-10}          &       &       & {Deep-S-MAD}   & \textbf{38.26\%} & 100.00\% & 100.00\% & 100.00\% & 0.00\% & 0.00\% \\
\cline{4-10}          &       &       & S-MBLBP   & 44.52\% & 100.00\% & 100.00\% & 100.00\% & 0.09\% & 0.00\% \\
    \hline
    \multicolumn{10}{|c|}{} \\
    \hline
    \multirow{4}[8]{*}{0.5} & \multirow{4}[8]{*}{1096} & \multirow{4}[8]{*}{1850} & PRNU   & 47.29\% & \textbf{85.86\%} & \textbf{97.45\%} & 100.00\% & 0.09\% & 0.00\% \\
\cline{4-10}          &       &       & SSE  & 53.74\% & 94.80\% & 97.99\% & \textbf{99.91\%} & 0.00\% & 0.00\% \\
\cline{4-10}          &       &       & {Deep-S-MAD}   & \textbf{35.43\%} & 100.00\% & 100.00\% & 100.00\% & 0.00\% & 0.00\% \\
\cline{4-10}          &       &       & S-MBLBP   & 42.15\% & 100.00\% & 100.00\% & 100.00\% & 0.09\% & 0.00\% \\
    \hline
    \end{tabular}%
    }
  \label{tab:S-MAD_P&S}%
\end{table*}%

\begin{figure}[!ht]
	\centering
	\includegraphics[width=0.85\textwidth]{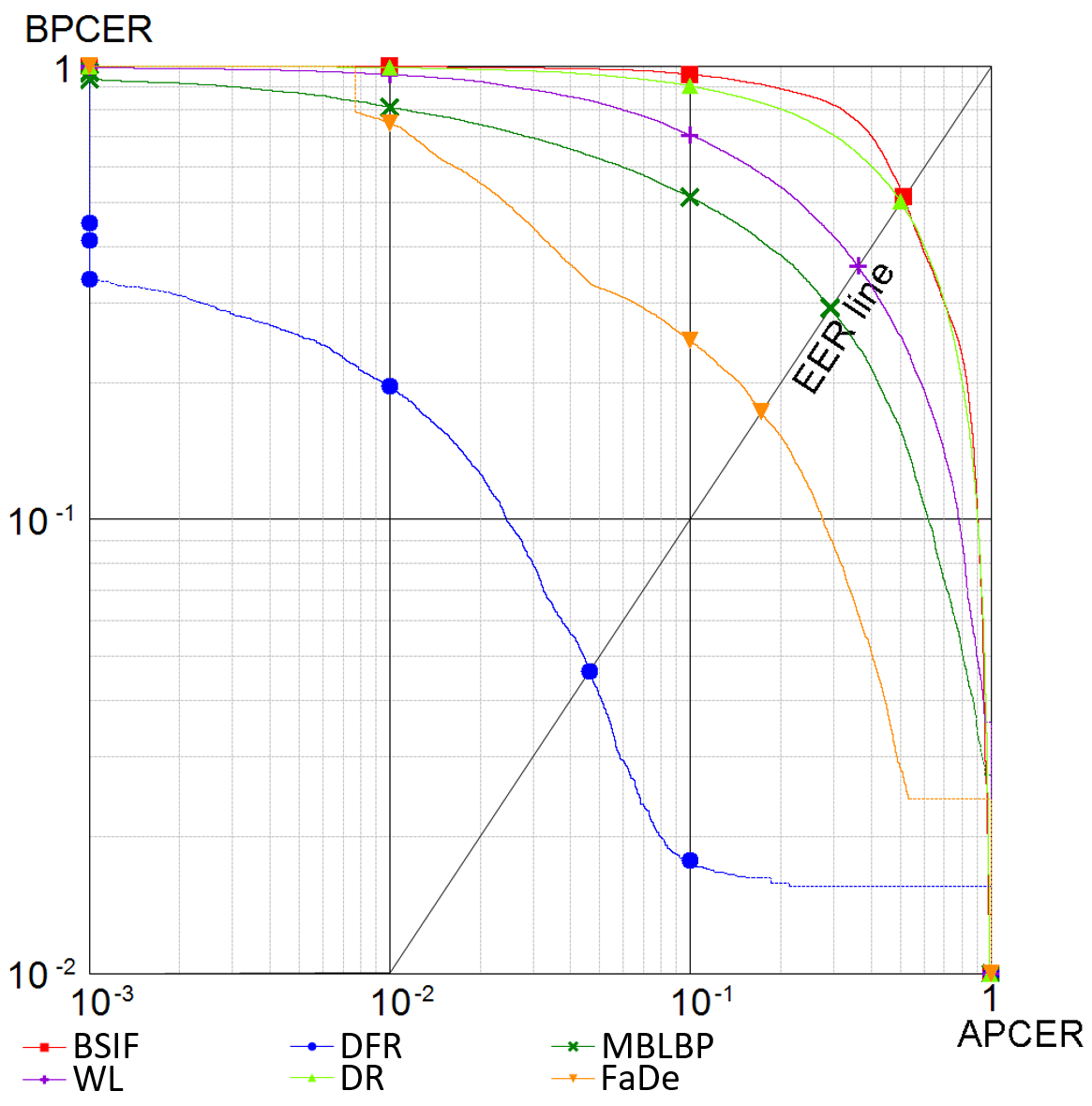}
	\hfill 
	\caption{DET plot for the D-MAD-SOTAMD\_P\&S-1.0.}
	\label{fig:DETPlotOverall_PS}
\end{figure}

The results obtained on the \textit{P\&S Image Benchmark (D-MAD-SOTAMD\_P\&S-1.0)} are summarized in Fig. \ref{fig:DETPlotOverall_PS}.
While for the best performing approach (DFR) the detection accuracy on Digital and P\&S images is similar, in general a performance drop on Print and Scan images can be observed; for example, for the demorphing method (FaDe) the BPCER values are about 10\% higher. Also in this case the influence of the morphing factor on the MAD performance can be observed in Table \ref{tab:overallRes_PS} reporting the results for the overall  set  of images and for the subsets of images with morphing factor 0.3 and 0.5. 
\subsection{Results - S-MAD}
\label{ssec:results-s-mad}
The results of S-MAD algorithms on printed-scanned images are given in Table \ref{tab:S-MAD_P&S} and on digital images in Table \ref{tab:S-MAD_Digital} (Appendix) respectively. In this case the overall performance is quite unsatisfactory in general and very far from the accuracy needed in real operational conditions. No significant differences can be observed between the different test cases: morphing factor 0.3 or 0.5, digital or printed-scanned images. We can conclude that morphing attack detection based on the analysis of the single image is still very complex, particularly in the presence of heterogeneous image sources, different processing pipelines and high quality morphs obtained through a careful selection of subjects and an accurate post-processing aimed at removing all visible artifacts. The results confirm again the importance of cross-database training and testing to improve the robustness of detection algorithms.
\subsection{Directions for Future Works}
\label{ssec:directions-for-future}

As noted from the results reported in the previous sections, it is evident that the accuracy of MAD does not meet the operational requirements. If we focus on BPCER\textsubscript{100}, we can see from Tables ~\ref{tab:overallRes_D} and \ref{tab:overallRes_PS} that the result is around 20\% for the best performing D-MAD approach. For all S-MAD algorithms (see Table~\ref{tab:S-MAD_P&S} and Table \ref{tab:S-MAD_Digital} in Appendix), BPCER\textsubscript{100} is higher than 90\%. From  a practical point of view, this behaviour would cause a considerable number of false alarms \Rev{and, as a consequence, a high number or false rejections during face verification at ABC gates. This would be unacceptable if we consider that operational face verification systems for ABC gates are expected to work at a False Accept Rate (FAR) of 0.1 per cent with a False Rejection Rate (FRR) not higher than 5\% \cite{frontex2015best}}.

\begin{itemize}[leftmargin=*]
    \item Given the number of covariates impacting the MAD performance such as age, gender and ethnicity, \Rev{accurate and better} algorithms need to be developed to address the complex challenge of morphing attacks. \Rev{The results presented in this work also suggest that the combination of approaches of different nature could lead to a general performance improvement.}
    \item  As it can also be noted from the Table~\ref{tab:overallRes_PS} that the print and scan process reduces the MAD accuracy to a larger extent. \Rev{Reliable and accurate}  algorithms need to be developed to improve the accuracy of the algorithms for detecting morphing attacks specifically when images are processed through the print and scan pipeline. 
    \item As a complementary direction, the human detection performance should be studied in a standardized manner to understand the key factors in spotting the morphing attacks on FRS.
\end{itemize}

\section{Conclusion and Summary}
\label{sec:conclusion}
Given the complex nature of the morphing attack detection and the impact on operational FRS, we presented a new evaluation framework and a new database of morphed images in this work. The sequesterd morphed dataset being publicly available allows researchers to benchmark their algorithms in a continuous manner to contribute to development of morphing attack detection. Further, this work also provides a benchmark of the existing state of the art algorithms to give a clear idea of the limitations in the existing algorithms for MAD.

\ifCLASSOPTIONcompsoc
  \section*{Acknowledgments}
\else
  \section*{Acknowledgment}
\fi
The authors would like to thank European Commission for supporting this work funded by SOTAMD project. The content of this report represents the views of the authors only and is their sole responsibility. The European Commission does not accept any responsibility for use that may be made of the information it contains. Further we are grateful to our colleagues at the German Federal Office for Information Security (BSI), the Hochschule Bonn-Rhein-Sieg (H-BRS) and to the Norwegian Police for the support in the data acquisition.

\bibliographystyle{IEEEtran}
\bibliography{sotamd-journal-tifs-accepted-200919}
\newpage

\begin{IEEEbiography}
		[{\includegraphics[width=1in,height=1.25in,clip,keepaspectratio]{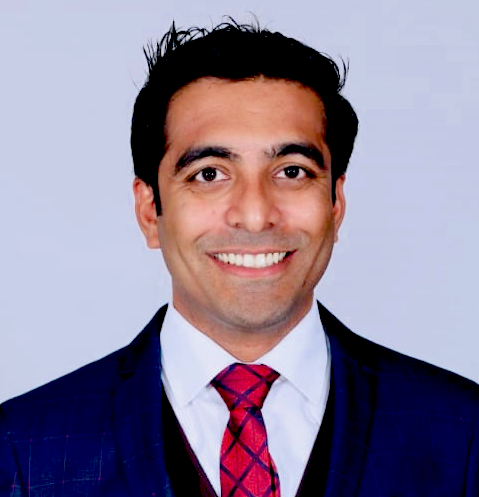}}]{Kiran Raja} obtained his PhD in computer Science from Norwegian University of Science and Technology (NTNU), Norway in 2016. He is faculty member at Dept. of Computer Science at NTNU, Norway. His main research interests include statistical pattern recognition, image processing, and machine learning with applications to biometrics, security and privacy protection. He was/is participating in EU projects SOTAMD, iMARS and other national projects. He has authored several papers in his field of interest and serves as a reviewer for number of journals and conferences. He is a member of EAB and chairs Academic Special Interest Group at EAB.
\end{IEEEbiography}
\vskip -3\baselineskip plus -1fil
\begin{IEEEbiography}
		[{\includegraphics[width=1in,height=1.25in,clip,keepaspectratio]{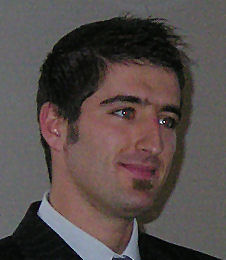}}]{Matteo Ferrara} is an associate professor at the Department of Computer Science and Engineering of the University of Bologna, Italy. His research interests are in the areas of Pattern Recognition, Computer Vision, Image Processing and Machine Learning. He received his bachelor's degree cum laude in Computer Science from the University of Bologna in 2004, his Master's degree cum laude in 2005 and his Ph.D. in 2009. Most of his applied research is in the field of Biometric Systems. He is member of the Biometric System Laboratory and he is one of the organizers of the international performance evaluation initiative named “FVC-onGoing”. Moreover, he is one of the authors of the well-known fingerprint recognition algorithm named “Minutia Cylinder-Code” (MCC). Finally he first proved that face morphing can be exploited to fool Automated Border Control (ABC) systems. He is author of several scientific papers and he served as referee for international conferences and journals. He took part to national and European research projects and to consultancy projects between the University of Bologna and foreign universities and companies.
\end{IEEEbiography}
\vskip -3\baselineskip plus -1fil
\begin{IEEEbiography}
		[{\includegraphics[width=1in,height=1.25in,clip,keepaspectratio]{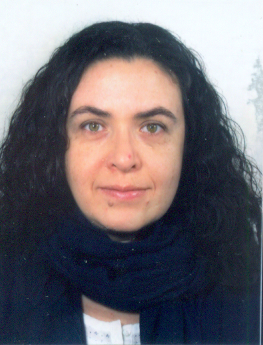}}]{Annalisa Franco} is Assistant Professor at the Department of Computer Science and Engineering, University of Bologna, Italy. In 2004 she received her Ph.D. in Electronics, Computer Science and Telecommunications Engineering at DEIS, University of Bologna for her work on Multidimensional Indexing Structures and their application is pattern recognition. She is a member of the Biometric System Laboratory at Computer Science - Cesena. She authored several scientific papers and served as referee for a number of international journals and conferences. Her research interests include Pattern Recognition, Biometric Systems, Image Databases and Multidimensional Data Structures. Recent research activity is mainly focused on automatic face recognition.
\end{IEEEbiography}
\vskip -3\baselineskip plus -1fil
\begin{IEEEbiography}
		[{\includegraphics[width=1in,height=1.25in,clip,keepaspectratio]{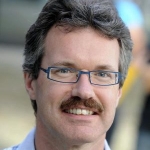}}]{Luuk Spreeuwers} studied Electrical Engineering at the University of Twente, Netherlands. In 1992 he obtained his PhD from the University of Twente. The title of his PhD-thesis is: Image Filtering with Neural Networks: Applications and Performance Evaluation. Subsequently Luuk Spreeuwers worked at the International Institute for Aerospace and Earth Sciences (ITC) in Enschede, Netherlands, the University of Twente in a SION project on 3-D image analysis of aerial image sequences and in Budapest at the Hungarian Academy of Sciences in a 3-D textures ERCIM project. From 1999-2005 Luuk Spreeuwers worked on 3-D modeling and segmentation of the human heart in MRI at the Image Sciences Institute of the University Medical Centre in Utrecht, the Netherlands. Currently, Luuk is an Associate Professor at the University of Twente and manages and is involved in various biometric research projects in the area of 2D and 3D face recognition, face morphing attack detection and finger vein recognition.
\end{IEEEbiography}
\vskip -3\baselineskip plus -1fil
\begin{IEEEbiography}
		[{\includegraphics[width=1in,height=1.25in,clip,keepaspectratio]{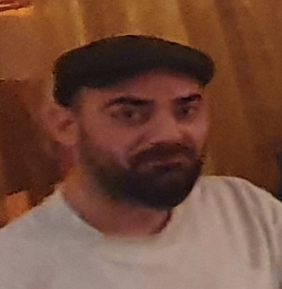}}]{Ilias Batskos} holds a diploma in Electrical and Computer Engineering from the Aristotle University of Thessaloniki and a MSc degree in Forensic Science from the University of Amsterdam. He is currently working towards a PhD in Computer Science at the University of Twente. At the Database Management and Biometrics (DMB) group he has focused his research on evaluation of biometric systems and the effect of morphing on biometric system performance.
\end{IEEEbiography}
\vskip -3\baselineskip plus -1fil
\begin{IEEEbiography}
		[{\includegraphics[width=1in,height=1.25in,clip,keepaspectratio]{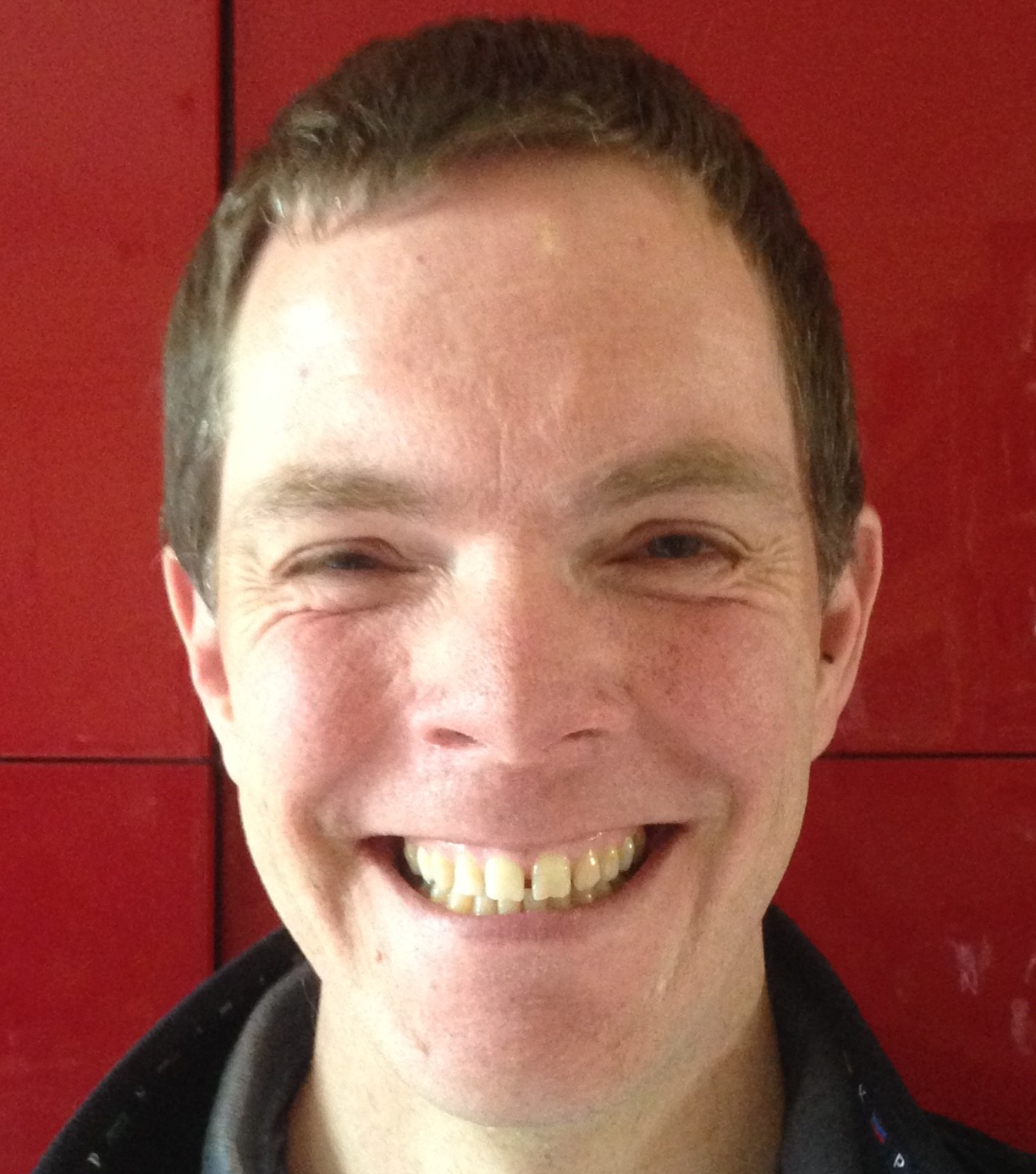}}]{Florens de Wit} is a researcher in biometrics at the University of Twente in the Netherlands. He received a MSc in Applied Physics in 1999 at the Eindhoven University of Technology, and in Forensic Science in 2006 at the University of Amsterdam. Before joining the Database Management and Biometrics (DMB) group of the faculty of Electrical Engineering, Mathematics and Computer Science (EEMCS) in 2017, he worked as a researcher at the Netherlands Forensic Institute (NFI) and as a lecturer at the Saxion University of Applied Science. At the DMB group he has focused his research on evaluation of biometric systems and the effect of morphing on biometric system performance.
\end{IEEEbiography}
\vskip -3\baselineskip plus -1fil
\begin{IEEEbiography}
		[{\includegraphics[width=1in,height=1.25in,clip,keepaspectratio]{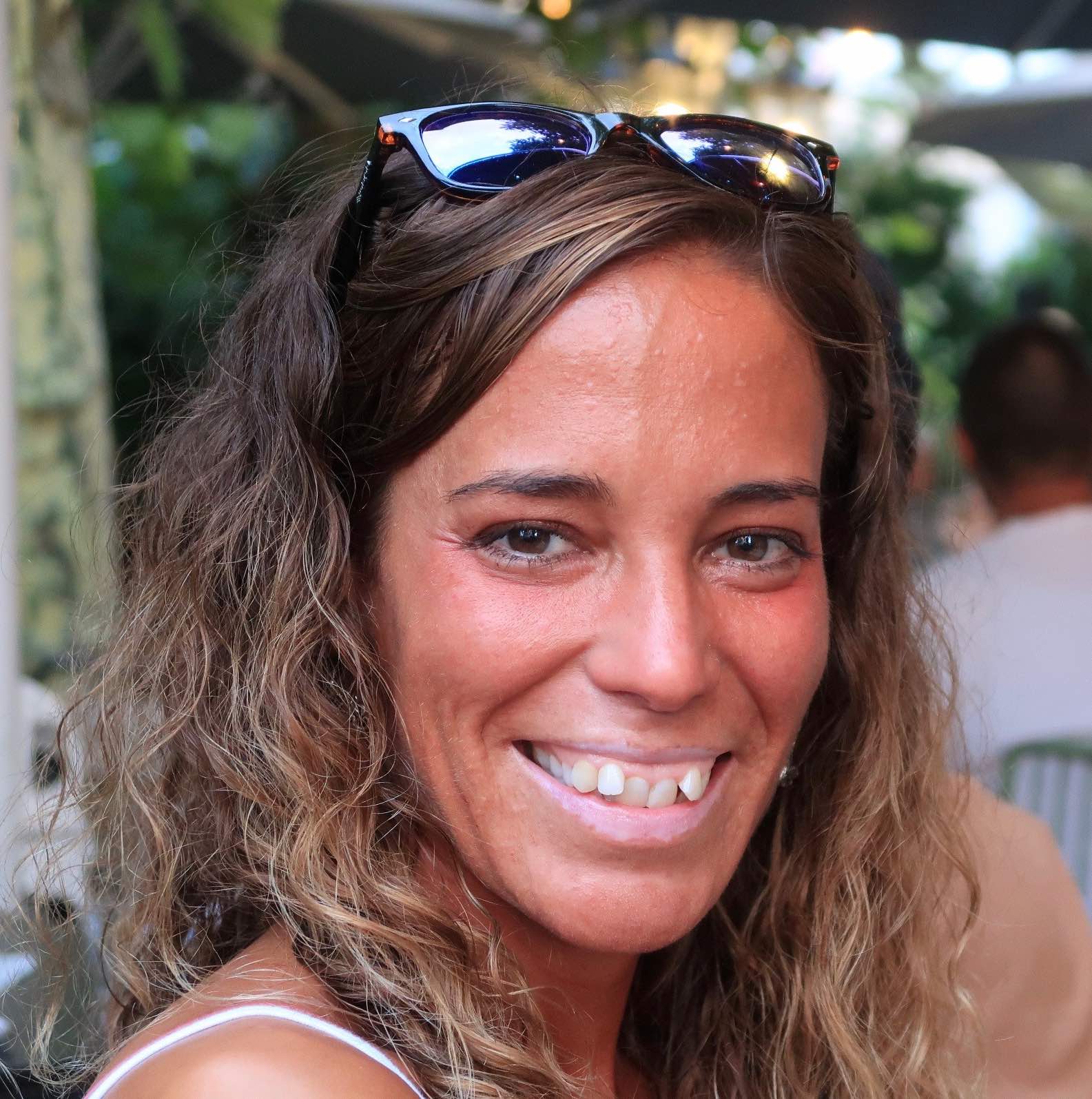}}]{ Marta Gomez-Barrero} is a Professor for IT-Security and technical data privacy at the Hochschule Ansbach, in Germany. Between 2016 and 2020, she was a postdoctoral researcher at the National Research Center for Applied Cybersecurity (ATHENE) - Hochschule Darmstadt, Germany. Before that, she received her MSc degrees in Computer Science and Mathematics (2011), and her PhD degree in Electrical Engineering (2016), all from Universidad Autonoma de Madrid, Spain. Her current research focuses on security and privacy evaluations of biometric systems, Presentation Attack Detection (PAD) methodologies, and biometric template protection (BTP) schemes. She has co-authored more than 70 publications, chaired special sessions and competitions at international conferences, she is associate editor for the EURASIP Journal on Information Security, and represents the German Institute for Standardization (DIN) in ISO/IEC SC37 JTC1 SC37 on biometrics. She has also received a number of distinctions, including: EAB European Biometric Industry Award 2015, Best Ph.D. Thesis Award by Universidad Autonoma de Madrid 2015/16, Siew-Sngiem Best Paper Award at ICB 2015, Archimedes Award for young researches from Spanish MECD, and Best Poster Award at ICB 2013.
\end{IEEEbiography}
\vskip -3\baselineskip plus -1fil
\begin{IEEEbiography}
		[{\includegraphics[width=1in,height=1.25in,clip,keepaspectratio]{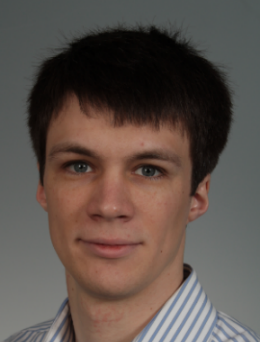}}]{Ulrich Scherhag} received his B.Eng degree (Electrical Engineering) in 2012 from the Duale Hochschule Baden-Württemberg, Mannheim. He stated studying computer science in 2014 at Hochschule Darmstadt and received the M.Sc degree (Computer Science, IT-Security) in 2016, for which he was granted the CAST Award IT-Security 2016. Since 2016 he is a Ph.D. Student Member of da/sec at the National Research Center for Applied Cybersecurity (ATHENE). He is a member of the European Association for Biometrics (EAB) and a Reviewer for the International Conference of the Biometrics Special Interest Group (BIOSIG) and IEEE Access. His current research focuses on presentation attack detection and morphed face detection.
\end{IEEEbiography}
\vskip -3\baselineskip plus -1fil
\begin{IEEEbiography}
		[{\includegraphics[width=1in,height=1.25in,clip,keepaspectratio]{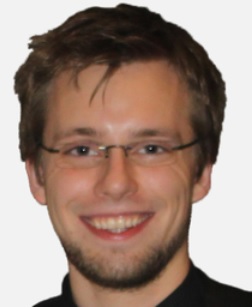}}]{Daniel Fischer}  completed his bachelor study program at University of Applied Sciences Darmstadt in cooperation with Deutsche Telekom AG in 2015, where he received his “Bachelor of Science” degree in Computer Science. He subsequently completed his master degree study program (“Master of Science” in Computer Science) with a focus on IT-Security and Biometrics at University of Applied Sciences Darmstadt in 2018. Since 2015 he is a member of the IT-Security group (“da/sec”) at University of Applied Sciences Darmstadt, which is a part of the National Research Center for Applied Cybersecurity (ATHENE).
\end{IEEEbiography}
\vskip -3\baselineskip plus -1fil
\begin{IEEEbiography}
		[{\includegraphics[width=1in,height=1.25in,clip,keepaspectratio]{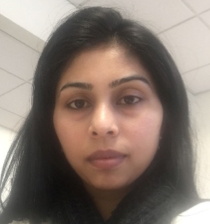}}]{Sushma Krupa Venkatesh} is a PhD candidate at Norwegian University of Science and Technology(NTNU), Norway since 2016. She obtained her bachelor’s degree in computer science in 2008 and master’s degree in computer science and technology in 2011. Her recent research interests include statistical pattern recognition, image processing, and machine learning with applications to biometrics, privacy and security. She has authored number of technical papers in various journals and conferences and serves as a reviewer for various scientific publication venues.
\end{IEEEbiography}
\vskip -3\baselineskip plus -1fil
\begin{IEEEbiography}
		[{\includegraphics[width=1in,height=1.25in,clip,keepaspectratio]{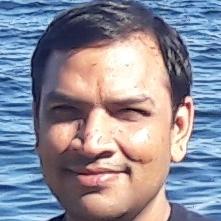}}]{Jag Mohan Singh} received the BTech (Honors) and MS degrees by research in computer science from the International Institute of Information Technology (IIIT), Hyderabad, in 2005 and 2008, respectively. He worked in the industrial research \& development departments of Intel, Samsung, Qualcomm, and Applied Materials in India from 2010 till 2018. He is currently PhD student at the Norwegian University of Science and Technology (NTNU Gjøvik) in Norwegian Biometrics Laboratory (NBL). His research interests include presentation attack detection for face and finger modalities.
\end{IEEEbiography}
\vskip -3\baselineskip plus -1fil
\begin{IEEEbiography}
		[{\includegraphics[width=1in,height=1.25in,clip,keepaspectratio]{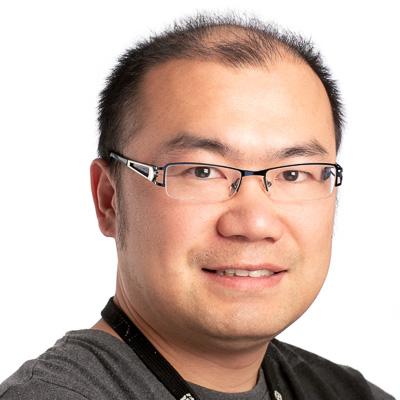}}]{Guoqiang Li} received his bachelor’s degree from JiLin University (2005, China), the master’s degree from Harbin Institute of Technology (2007, China), and his PhD degree in information security from Norwegian University of Technology and Science in 2016. He is currently working as a senior researcher in the Department of Information Security and Communication Technology at Norwegian University of Science and Technology (NTNU), Gjøvik, Norway. His research interests are related to fingerprint recognition, face recognition, biometric template protection, behavioural biometrics, and image processing.
\end{IEEEbiography}
\vskip -3\baselineskip plus -1fil
\begin{IEEEbiography}
		[{\includegraphics[width=1in,height=1.25in,clip,keepaspectratio]{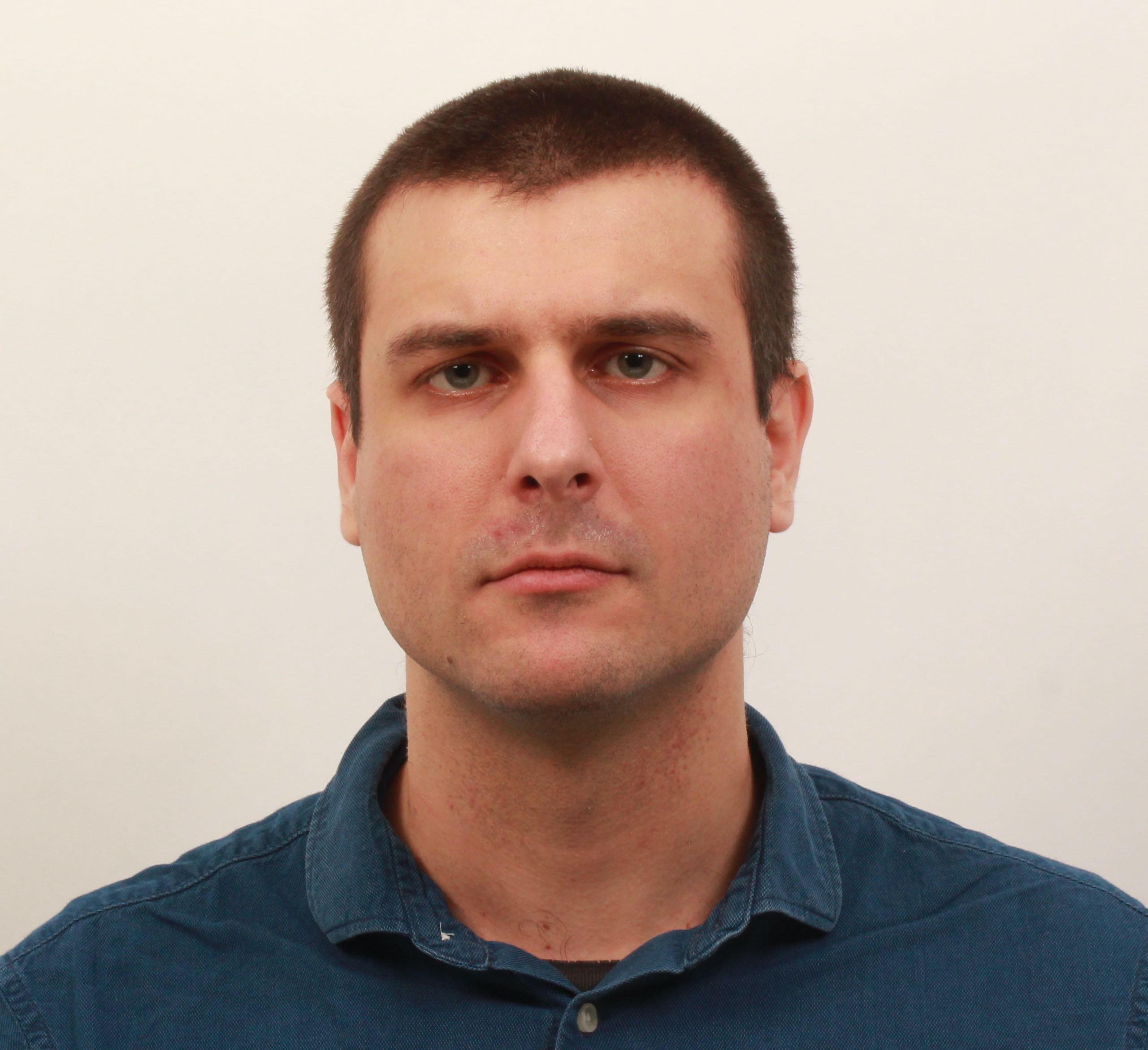}}]{Loïc Bergeron} obtained his Master's degree in Computer security in 2018 at Caen Normandie University. After his participation in a research project on keystrokes dynamics at the Norwegian University of Science and Technology (NTNU), he became a member of the SOTAMD project in 2019. He was involved in the different parts of this biometric research project and he mainly coordinated the data collection.
\end{IEEEbiography}
\vskip -3\baselineskip plus -1fil
\begin{IEEEbiography}
		[{\includegraphics[width=1in,height=1.25in,clip,keepaspectratio]{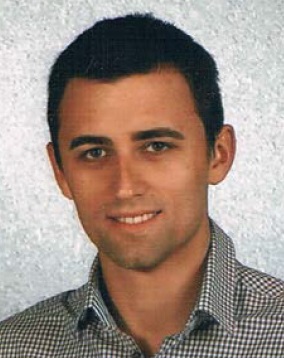}}]{Sergey Isadskiy} holds a Master degree in computer science from the Hochschule Darmstadt, Germany. He is a former member of the da/sec - Biometrics and Internet Security research group at Hochschule Darmstadt and the National Research Center for Applied Cybersecurity (ATHENE). His research interests include pattern recognition with focus on biometrics, in particular face and speaker recognition.
\end{IEEEbiography}
\vskip -3\baselineskip plus -1fil
\begin{IEEEbiography}
	[{\includegraphics[width=1in,height=1.25in,clip,keepaspectratio]{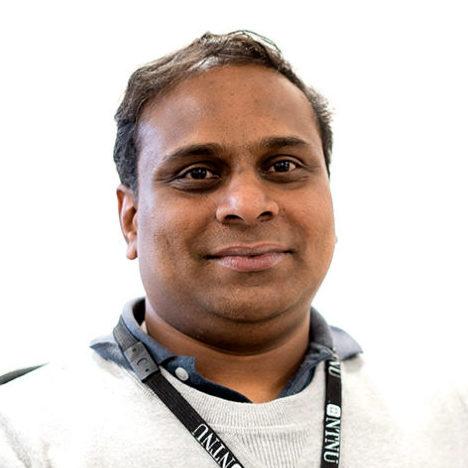}}]{Raghavendra Ramachandra} received the bachelor’s degree from the University of Mysore (UOM), Mysore, India, the master’s degree in electronics and communication from Visvesvaraya Technological University, Belgaum, India, and the Ph.D. degree in computer science and technology from UOM and Institute Telecom, and Telecom Sudparis, Évry, France (carried out as a collaborative work). He is currently appointed as a Professor with the Norwegian Biometric Laboratory, Norwegian University of Science and Technology (NTNU), Gjøvik, Norway. He was a Researcher with the Istituto Italiano di Tecnologia, Genoa, Italy. His main research interests include statistical pattern recognition, data fusion schemes and random optimization, with applications to biometrics, multimodal biometric fusion, human behavior analysis, and crowd behavior analysis. He has authored several papers, and is a reviewer for several international conferences and journals. He also involved in various conference organising and program committees. He is also an associate editor for various journals.
\end{IEEEbiography}
\vskip -3\baselineskip plus -1fil
\begin{IEEEbiography}
		[{\includegraphics[width=1in,height=1.25in,clip,keepaspectratio]{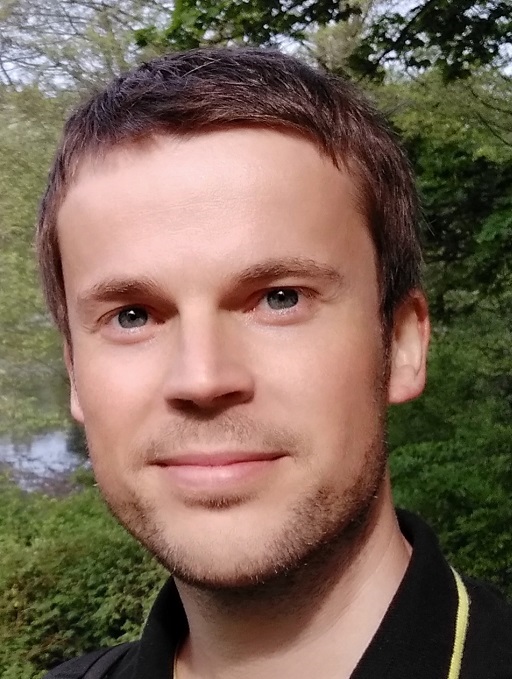}}]{Christian Rathgeb} is a Senior Researcher with the Faculty of Computer Science, Hochschule Darmstadt (HDA), Germany. He is a Principal Investigator in the National Research Center for Applied Cybersecurity (ATHENE). His research includes pattern recognition, iris and face recognition, security aspects of biometric systems, secure process design and privacy enhancing technologies for biometric systems. He co-authored over 100 technical papers in the field of biometrics. He is a winner of the EAB - European Biometrics Research Award 2012, the Austrian Award of Excellence 2012, Best Poster Paper Awards (IJCB'11, IJCB'14, ICB'15) and the Best Paper Award Bronze (ICB'18). He is a member of the European Association for Biometrics (EAB), a Program Chair of the International Conference of the Biometrics Special Interest Group (BIOSIG) and a editorial board member of IET Biometrics (IET BMT). He has served for various program committees and conferences (e.g. ICB, IJCB, BIOSIG, IWBF) and journals  as a reviewer (e.g. IEEE TIFS, IEEE TBIOM, IET BMT).
\end{IEEEbiography}
\vskip -3\baselineskip plus -1fil
\begin{IEEEbiography}
		[{\includegraphics[width=1in,height=1.25in,clip,keepaspectratio]{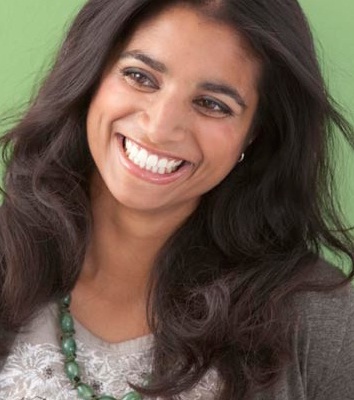}}]{Dinusha Frings} is specialised in managing projects related to biometrics, digital identity and travel documents. She currently holds two positions. One at the National Office for Identity Data (NOID) and one at the European Association for Biometrics (EAB).She coordinated the State Of The Art of Morphing Detection (SOTAMD), Known Traveller Digital Identity and Research Live Enrolment on behalf of NOID. And is, on behalf of the EAB, currently involved in project iMARS, short for image Manipulation Attack Resolving Solutions. She chairs of the EAB Operator Special Interest Group and is part of the organising committee of the EAB Research Project Conference. Dinusha previously worked at IDEMIA in the Government Identity Solutions division and coordinated multiple IT projects related to secure ID documents.
\end{IEEEbiography}
\vskip -3\baselineskip plus -1fil
\begin{IEEEbiography}
		[{\includegraphics[width=1in,height=1.25in,clip,keepaspectratio]{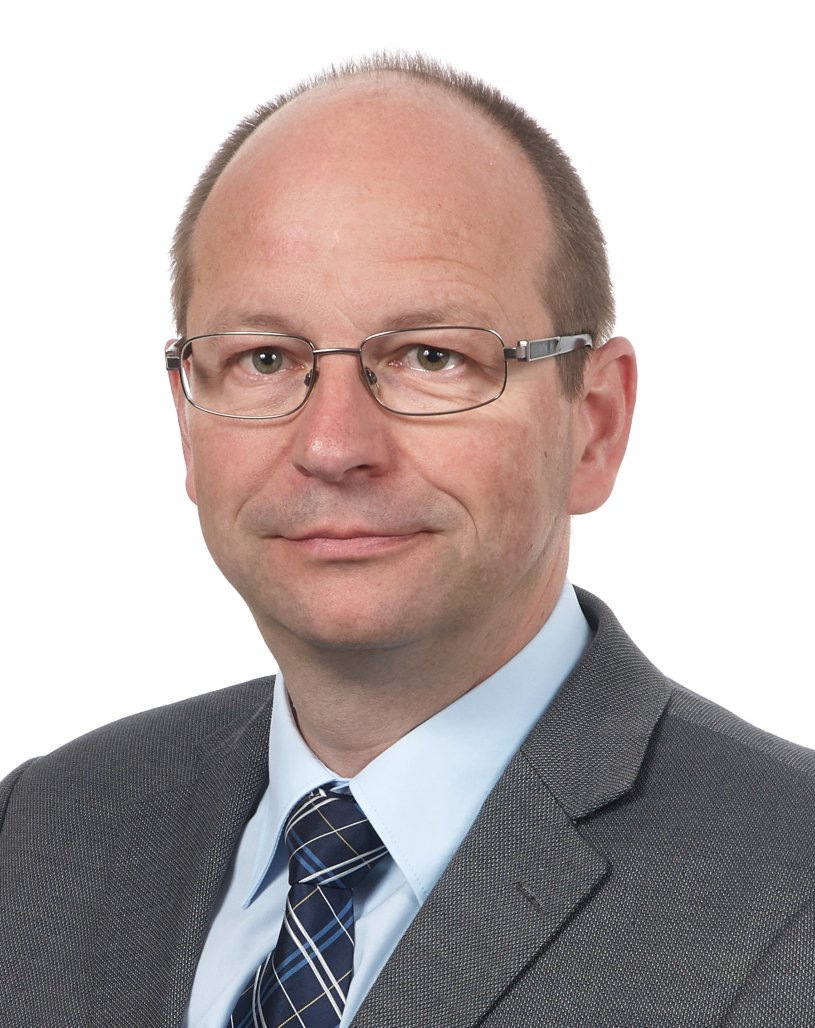}}]{Uwe Seidel} received his Ph.D. degree in experimental physics and optics from the University of Jena. After some years in industry, he joined the Forensic Science Institute of the German Federal Criminal Police Office (Bundeskriminalamt) as an ID document expert in 2000 and is now heading the BKA’s IT Forensics and Document section. This assignment also involves overseeing R\&D projects for increasing the counterfeit resistance of German official documents. Since 2019, he is the chairman of ICAO's New Technology Working Group and Germany’s Alternate Member for the ICAO Technical Advisory Group TAG/TRIP.
\end{IEEEbiography}
\vskip -3\baselineskip plus -1fil
\begin{IEEEbiography}
		[{\includegraphics[width=1in,height=1.25in,clip,keepaspectratio]{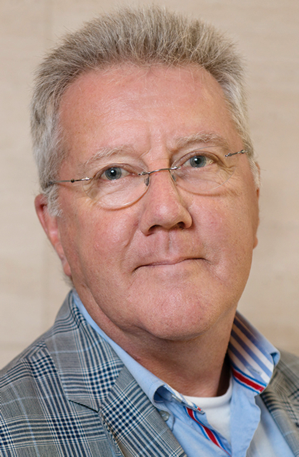}}]{Fons Knopjes} is Senior Advisor at the National Office for Identity Data (Dutch Ministry of the Interior and Kingdomrelations). He has advised different Governments during the development of many (electronic)  passports and identity documents. His last advice was about Dutch travel documents introduced in 2014.  He redesigned the identification process of suspects arrested by the police and has advised Governments by the restructuring of their identity infrastructure. Fons has developed a model aimed at the dynamic management of the security concept of documents. He executed assessments on the identity infrastructure in over 10 countries in Europe, Africa and Asia. High level managers of over 35 countries of all over de globe have participated in Master Classes that Fons has given. He participated in the EU-funded SOTAMD research project. Fons is a member of the core group of experts on Identity related crime of the UNODC (United Nations Office on Drugs and Crime), ICAO’s TAG-TRIP (Traveler Identification Program), IAI (International Association for Identification) and the board of advisors Center for Identity, University of Texas (US).
\end{IEEEbiography}
\vskip -3\baselineskip plus -1fil
\begin{IEEEbiography}
		[{\includegraphics[width=1in,height=1.25in,clip,keepaspectratio]{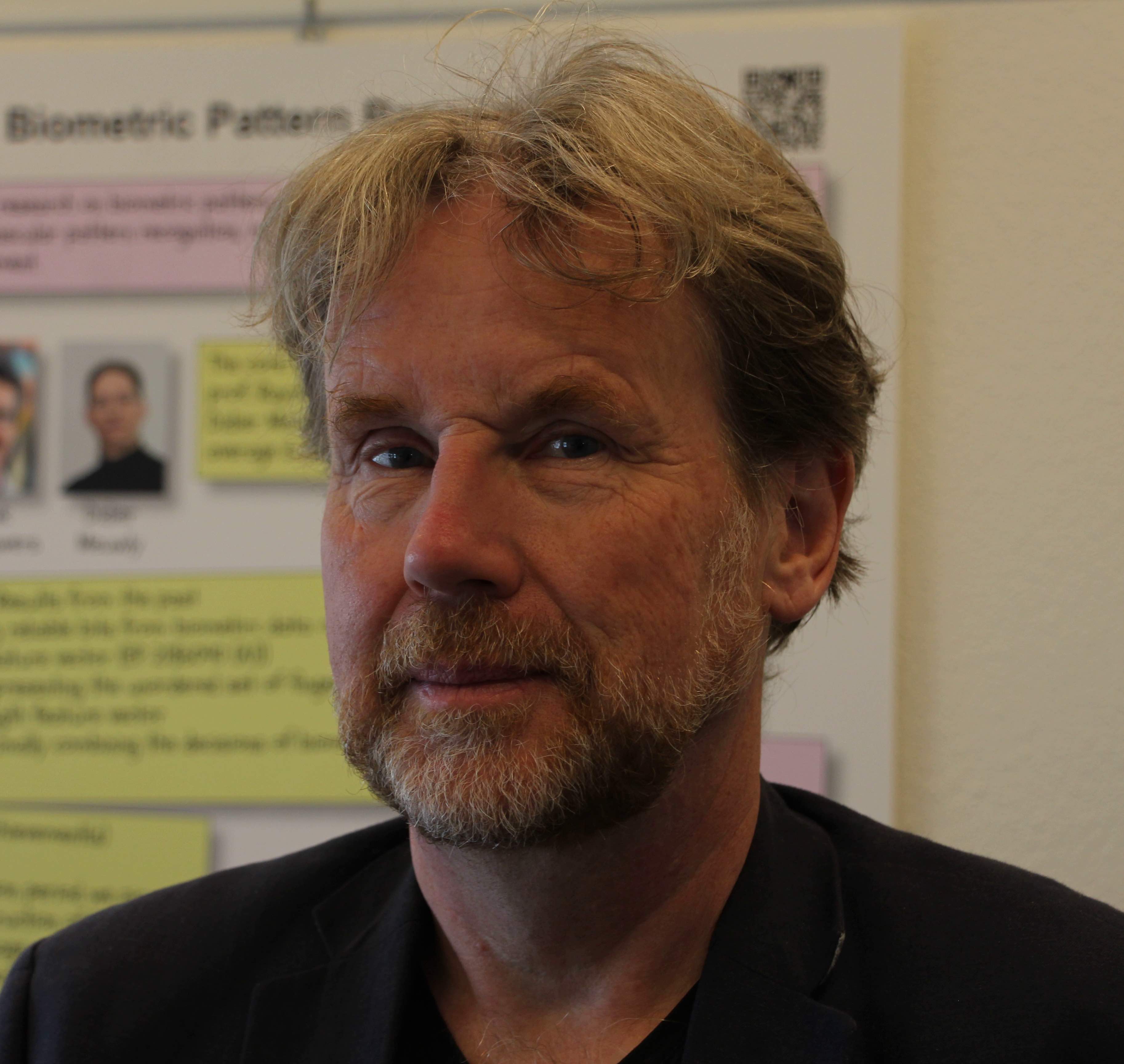}}]{Raymond Veldhuis} graduated from the University of Twente, The Netherlands, in 1981. He received the Ph.D. degree from Nijmegen University on a thesis entitled Adaptive Restoration of Lost Samples in Discrete-Time Signals and Digital Images, in 1988. From 1982 to 1992, he was a Researcher with Philips Research Laboratories, Eindhoven, in various areas of digital signal processing. From 1992 to 2001, he was involved in the field of speech processing. He is currently a Full Professor in Biometric Pattern Recognition with the University of Twente, where he is leading the Data Management and Biometrics group. His main research topics are face recognition (2-D and 3-D), fingerprint recognition, vascular pattern recognition, multibiometric fusion, and biometric template protection. The research is both applied and fundamental. 
\end{IEEEbiography}
\vskip -3\baselineskip plus -1fil
\begin{IEEEbiography}
		[{\includegraphics[width=1in,height=1.25in,clip,keepaspectratio]{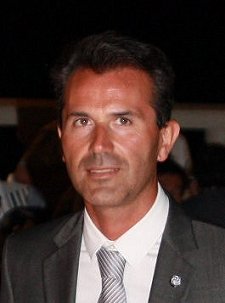}}]{Davide Maltoni} is a Full Professor at University of Bologna (Dept. of Computer Science and Engineering - DISI). His research interests are in the area of Pattern Recognition, Computer Vision, Machine Learning and Computational Neuroscience. Davide Maltoni is co-director of the Biometric Systems Laboratory (BioLab), which is internationally known for its research and publications in the field. Several original techniques have been proposed by BioLab team for fingerprint feature extraction, matching and classification, for hand shape verification, for face location and for performance evaluation of biometric systems. Davide Maltoni is co-author of the Handbook of Fingerprint Recognition published by Springer, 2009 and holds three patents on Fingerprint Recognition. He has been elected IAPR (International Association for Pattern Recognition) Fellow 2010.
\end{IEEEbiography}
\vskip -3\baselineskip plus -1fil
\begin{IEEEbiography}
		[{\includegraphics[width=1in,height=1.25in,clip,keepaspectratio]{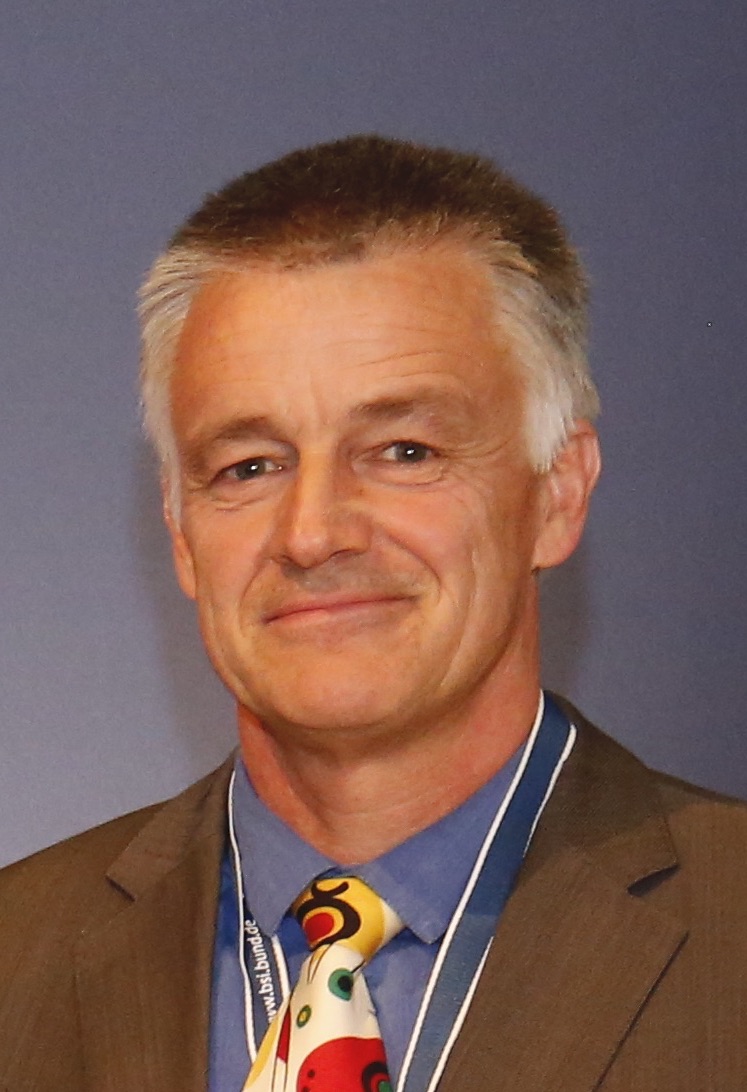}}]{Christoph Busch} is member of the Norwegian University of Science and Technology (NTNU), Norway. He holds a joint appointment with Hochschule Darmstadt (HDA), Germany. Further he lectures Biometric Systems at Denmark’s DTU since 2007. On behalf of the German BSI he has been the coordinator for the project series BioIS, BioFace, BioFinger, BioKeyS Pilot-DB, KBEinweg and NFIQ2.0. He was//is partner of the EU projects 3D-Face, FIDELITY, TURBINE, SOTAMD, RESPECT, TReSPsS, iMARS and others. He is also principal investigator in the German National Research Center for Applied Cybersecurity (ATHENE) and is co-founder of the European Association for Biometrics (EAB). Christoph co-authored more than 500 technical papers and has been a speaker at international conferences. He is member of the editorial board of the IET journal on Biometrics and of IEEE TIFS journal. Furthermore he chairs the TeleTrusT biometrics working group as well as the German standardization body on Biometrics and is convenor of WG3 in ISO/IEC JTC1 SC37.
\end{IEEEbiography}

%
\ifCLASSOPTIONcaptionsoff
  \newpage
\fi

\newpage
\counterwithin{figure}{section}
\counterwithin{table}{section}
\onecolumn
\begin{appendices}

\section{Additional Analysis - D-MAD Results on overall set of images and for the subsets of images with morphing factor 0.3 and 0.5}
\begin{table*}[htbp]
  \centering
  \caption{Performance indicators measured on the D-MAD-SOTAMD\_D-1.0 benchmark for the overall set of images and for the subsets of images with morphing factor 0.3 and 0.5.}
  \resizebox{0.9\textwidth}{!}{
    \begin{tabular}{|c|c|c|p{6.22em}|p{4.055em}|p{4.055em}|p{4.055em}|p{4.78em}|p{4.055em}|p{4.055em}|}
    \hline
    \rowcolor[rgb]{ .851,  .851,  .851} \multicolumn{1}{|p{4.055em}|}{\textbf{Test}} & \multicolumn{1}{p{6.055em}|}{\textbf{Bona fide comparisons}} & \multicolumn{1}{p{5.945em}|}{\textbf{Morphed comparisons}} & \textbf{Algorithm} & \textbf{EER} & \textbf{BPCER\textsubscript{10}} & \textbf{BPCER\textsubscript{20}} & \textbf{BPCER\textsubscript{100}} & \textbf{REJ\textsubscript{NBFRA}} & \textbf{REJ\textsubscript{NMRA}} \\
    \hline
    \multirow{5}[10]{*}{Overall} & \multirow{5}[10]{*}{3000} & \multirow{5}[10]{*}{30550} & BSIF & 45.93\% & 78.30\% & 84.13\% & 93.83\% & 1.53\% & 1.42\% \\
\cline{4-10}          &       &       & DFR & \textbf{4.54\%} & \textbf{2.00\%} & \textbf{3.93\%} & \textbf{18.87\%} & 1.67\% & 1.55\% \\
\cline{4-10}          &       &       & MBLBP & 33.47\% & 52.80\% & 59.93\% & 74.80\% & 2.80\% & 2.62\% \\
\cline{4-10}          &       &       & WL & 37.13\% & 71.67\% & 83.27\% & 95.67\% & 3.33\% & 3.16\% \\
\cline{4-10}          &       &       & DR  & 52.03\% & 89.70\% & 94.70\% & 98.57\% & \textbf{0.00\%} & \textbf{0.00\%} \\
\cline{4-10}          &       &       & FaDe   & 14.17\% & 17.20\% & 22.77\% & 64.57\% & 0.20\% & 0.19\% \\
    \hline
    \multicolumn{10}{|c|}{} \\
    \hline
    \multirow{5}[10]{*}{0.3} & \multirow{5}[10]{*}{3000} & \multirow{5}[10]{*}{10350} & BSIF & 46.43\% & 78.50\% & 85.23\% & 94.50\% & 1.53\% & 1.40\% \\
\cline{4-10}          &       &       & DFR & \textbf{1.96\%} & \textbf{1.67\%} & \textbf{1.67\%} & \textbf{13.23\%} & 1.67\% & 1.54\% \\
\cline{4-10}          &       &       & MBLBP & 31.57\% & 49.67\% & 56.47\% & 68.00\% & 2.80\% & 2.60\% \\
\cline{4-10}          &       &       & WL & 32.87\% & 63.90\% & 77.73\% & 93.97\% & 3.33\% & 3.07\% \\
\cline{4-10}          &       &       & DR  & 52.60\% & 90.00\% & 94.77\% & 98.57\% & \textbf{0.00\%} & \textbf{0.00\%} \\
\cline{4-10}          &       &       & FaDe   & 8.43\% & 7.27\% & 12.60\% & 27.47\% & 0.20\% & 0.17\% \\
    \hline
    \multicolumn{10}{|c|}{} \\
    \hline
    \multirow{5}[10]{*}{0.5} & \multirow{5}[10]{*}{3000} & \multirow{5}[10]{*}{20200} & BSIF & 45.70\% & 78.10\% & 83.53\% & 93.47\% & 1.53\% & 1.43\% \\
\cline{4-10}          &       &       &DFR & \textbf{5.40\%} & \textbf{2.47\%} & \textbf{5.60\%} & \textbf{21.50\%} & 1.67\% & 1.56\% \\
\cline{4-10}          &       &       & MBLBP & 34.33\% & 54.20\% & 61.93\% & 77.07\% & 2.80\% & 2.62\% \\
\cline{4-10}          &       &       & WL & 39.47\% & 74.47\% & 84.70\% & 96.40\% & 3.33\% & 3.20\% \\
\cline{4-10}          &       &       & DR  & 51.74\% & 89.60\% & 94.63\% & 98.57\% & \textbf{0.00\%} & \textbf{0.00\%} \\
\cline{4-10}          &       &       & FaDe   & 15.76\% & 20.07\% & 27.70\% & 100.00\% & 0.20\% & 0.19\% \\
    \hline
    \end{tabular}%
    }
  \label{tab:overallRes_D}%
\end{table*}%

\section{Additional Analysis - S-MAD Results on overall set of images and for the subsets of images with morphing factor 0.3 and 0.5}
\begin{table*}[htbp]
  \centering
  \caption{Performance indicators measured on the S-MAD-SOTAMD\_D-1.0 benchmark for the overall set of images and for the subsets of images with morphing factor 0.3 and 0.5.}
  \resizebox{0.9\textwidth}{!}{
    \begin{tabular}{|c|c|c|l|r|r|r|r|r|r|}
    \hline
    \rowcolor[rgb]{ .851.  .851.  .851} \multicolumn{1}{|p{4.055em}|}{\textbf{Test}} & \multicolumn{1}{p{6.055em}|}{\textbf{Bona fide comparisons}} & \multicolumn{1}{p{5.945em}|}{\textbf{Morphed comparisons}} & \textbf{Algorithm} & \textbf{EER} & \textbf{BPCER\textsubscript{10}} & \textbf{BPCER\textsubscript{20}} & \textbf{BPCER\textsubscript{100}} & \textbf{REJ\textsubscript{NBFRA}} & \textbf{REJ\textsubscript{NMRA}} \\
    \hline
    \multirow{4}[8]{*}{Overall} & \multirow{4}[8]{*}{300} & \multirow{4}[8]{*}{2045} & PRNU   & 44.81\% & 100.00\% & 100.00\% & 100.00\% & 0.00\% & 0.00\% \\
\cline{4-10}          &       &       & SSE  & \textbf{31.80\%} & \textbf{65.00\%} & \textbf{79.33\%} & \textbf{91.67\%} & 0.00\% & 0.00\% \\
\cline{4-10}          &       &       & {Deep-S-MAD}   & 38.99\% & 100.00\% & 100.00\% & 100.00\% & 0.00\% & 0.00\% \\
\cline{4-10}          &       &       & S-MBLBP   & 41.38\% & 100.00\% & 100.00\% & 100.00\% & 0.00\% & 0.00\% \\
    \hline
    \multicolumn{10}{|c|}{} \\
    \hline
    \multirow{4}[8]{*}{0.3} & \multirow{4}[8]{*}{300} & \multirow{4}[8]{*}{1035} & PRNU   & 44.81\% & 100.00\% & 100.00\% & 100.00\% & 0.00\% & 0.00\% \\
\cline{4-10}          &       &       & SSE  & \textbf{32.76\%} & \textbf{68.00\%} & \textbf{81.33\%} & \textbf{90.67\%} & 0.00\% & 0.00\% \\
\cline{4-10}          &       &       & {Deep-S-MAD}   & 39.64\% & 100.00\% & 100.00\% & 100.00\% & 0.00\% & 0.00\% \\
\cline{4-10}          &       &       & S-MBLBP   & 42.33\% & 100.00\% & 100.00\% & 100.00\% & 0.00\% & 0.00\% \\
    \hline
    \multicolumn{10}{|c|}{} \\
    \hline
    \multirow{4}[8]{*}{0.5} & \multirow{4}[8]{*}{300} & \multirow{4}[8]{*}{1010} & PRNU   & 44.82\% & 100.00\% & 100.00\% & 100.00\% & 0.00\% & 0.00\% \\
\cline{4-10}          &       &       & SSE  & \textbf{31.14\%} & \textbf{63.33\%} & \textbf{77.00\%} & \textbf{92.33\%} & 0.00\% & 0.00\% \\
\cline{4-10}          &       &       & {Deep-S-MAD}   & 38.33\% & 100.00\% & 100.00\% & 100.00\% & 0.00\% & 0.00\% \\
\cline{4-10}          &       &       & S-MBLBP   & 40.63\% & 100.00\% & 100.00\% & 100.00\% & 0.00\% & 0.00\% \\
    \hline
    \end{tabular}%
    }
  \label{tab:S-MAD_Digital}%
\end{table*}%

\newpage
\section{Additional Analysis - Attributes and subsets used for performance evaluation}
\begin{table}[htbp]
  \centering
  \caption{List of attributes and subsets used for performance evaluation.}
  \resizebox{0.58\textwidth}{!}{
    \begin{tabular}{|l|p{16.22em}|}
    \hline
    \rowcolor[rgb]{ .851,  .851,  .851} \multicolumn{1}{|c|}{\textbf{Attribute}} & \multicolumn{1}{c|}{\textbf{Attribute value}} \\
    \hline
    \multicolumn{1}{|l|}{\multirow{2}[4]{*}{Gender}} & Female \\
    \cline{2-2}          & Male \\
    \hline
    \multicolumn{1}{|l|}{\multirow{5}[10]{*}{Ethnicity}} & African \\
\cline{2-2}          & East-Asian \\
\cline{2-2}          & European/American \\
\cline{2-2}          & Indian-Asian \\
\cline{2-2}          & Middle Eastern \\
    \hline
    \multicolumn{1}{|l|}{\multirow{3}[6]{*}{Age}} & 18..35 \\
\cline{2-2}          & 36..55 \\
\cline{2-2}          & 56..75 \\
    \hline
    \multicolumn{1}{|l|}{\multirow{3}[6]{*}{Traits}} & Freckles \\
\cline{2-2}          & Moles \\
\cline{2-2}          & None – no relevant facial traits \\
    \hline
    \multicolumn{1}{|l|}{\multirow{3}[6]{*}{Partner}} & HDA \\
\cline{2-2}          & NTNU \\
\cline{2-2}          & UTW \\
    \hline
    \multirow{2}[4]{*}{Post-processing} & Automatic – no manual retouching \\
\cline{2-2}          & Manual – with manual retouching \\
    \hline
    Morphing algorithm & See \Rev{Table~\ref{tab:morphing-dataset-methos-processing} and Table~\ref{tab:morphing-dataset-acronyms}}\\
    \hline
    \multicolumn{1}{|p{11.61em}|}{Manual post-processing} & See \Rev{Table~\ref{tab:morphing-dataset-methos-processing} and Table~\ref{tab:morphing-dataset-acronyms}}\\
    \hline
    \Rev{Print-scan and } & \Rev{See  Table~\ref{tab:print-scan-details}}\\
     \Rev{and Compression} & \\
    \hline
    \multicolumn{1}{|l|}{\multirow{2}[4]{*}{Morph quality}} & Low – the morphed image is rejected at face verification stage by at least one FR SDK between Neurotechnology and Cognitec \\
\cline{2-2}          & High – the morphed image is accepted at face verification stage by both Neurotechnology and Cognitec FR SDKs \\
    \hline
    \end{tabular}%
    }
  \label{tab:test_subsets}%
\end{table}%

\begin{table}[htbp]
	\centering
	\resizebox{0.8\textwidth}{!}{
	\begin{tabular}{|c|c|c|c|c|c|}
		\hline
		 \textbf{Acronym} & \textbf{Algorithm} & \textbf{Acronym}& \textbf{Automated} & \textbf{Acronym} & \textbf{Manual }\\
		  & \textbf{description} & & \textbf{Post-Processing method} & & \textbf{post-processing method}\bigstrut\\
		\hline
		C01   & FaceMorpher \cite{FaceMorpher} & PA01  & Facemorpher's internal   & PM00  & No manual \bigstrut\\
		& & & post-processing + sharpening & & post-processing \bigstrut\\
		\hline
		C02   & FaceFusion  \cite{FaceFusion} & PA02  & FaceFusion's internal & PM01  & GIMP retouching \cite{GIMP} \\
		& (only used by HDA) & & post-processing+ sharpening & & \bigstrut\\
		\hline
		C03   & FaceMorph  \cite{dlib}  & PA03  & The replacement of the eye region  & PM02  & GIMP retouching \bigstrut\\
		& (OpenCV with Dlib) & &  is performed in post-processing, & &\bigstrut\\
		& & & to prevent a double iris. & &\bigstrut\\
		\hline
		C04   & FantaMorph \cite{Fantamorph} & PA04  & Fantamorph's & PM03  & Adobe Photoshop  \bigstrut\\
		& (only used by NTN)  & & internal processing & & retouching \cite{Photoshop}\bigstrut\\
		\hline
		C05   & Triangulation & PA05  & Background replacement & PM04  & Adobe Photoshop \bigstrut\\
		& with Dlib-landmarks & & , edge suppression, & & retouching\bigstrut\\
		&  & &  colour equalization & & \bigstrut\\
		\hline
		C06   & Triangulation  & PA06  & Background replacement,  & PM05  & GIMP retouching \bigstrut\\
		& with STASM-landmarks \cite{STASM} & & Poisson image editing \cite{10.1145/1201775.882269} & & \bigstrut\\
		\hline
		C07   & Triangulation  & PA07  & Background replacement, & PM06  & GIMP  \bigstrut\\
		& with NT-landmarks  & &  edge suppression,  & & retouching\bigstrut\\
		& & & colour equalization& &\bigstrut\\
		\hline
	\end{tabular}
	}
	\caption{Listing of various methods for morphing, automated and manual post-processing}
	\label{tab:morphing-dataset-acronyms}
\end{table}

	\begin{table}[htbp]
  \centering
  \caption{ Details of print-scan and compression pipeline along with image size}
  \resizebox{\textwidth}{!}{
      \begin{tabular}{|p{4.085em}|p{30.835em}|p{4.165em}|p{31em}|}
    \hline
    \multicolumn{1}{|c|}{\textbf{Acronym}} & \multicolumn{1}{c|}{\textbf{Print-Scan and Compression Pipeline}} & \multicolumn{1}{c|}{\textbf{Acronym}} & \multicolumn{1}{c|}{\textbf{Print-Scan and Compression Pipeline}} \bigstrut\\
    \hline
     F01 & 1.	Printed with printer Dmb DS-RX1HS at 300x300 dpi, matte\newline{}2.	Scanned using Epson V600 at 700 dpi\newline{}3.	Cropped to 785 x 1047 pixels\newline{}4.	Resized to 400 dpi 449x599 pixels\newline{}5.	Compressed using JPEG2000, max file size=15kb, RGB\_24\_BIT & F13 & 1.	Printed at professional photo laboratory of Fotogena \newline{}2.	Scanned using ID Document Scanner (Idemia)  \newline{} 3.Retrieved (compressed) image JPEG2000,max file size=15kb \bigstrut\\
    \hline
    F02 & 1.	Printed with printer Dmb DS-RX1HS at 300x600 dpi, matte\newline{}2.	Scanned using Epson V600 at 700 dpi\newline{}3.	Cropped to 785 x 1047 pixels\newline{}4.	Resized to 400 dpi 449x599 pixels\newline{}5.	Compressed using JPEG2000, max file size=15kb, RGB\_24\_BIT & F21 & 1. Created A4 PDFs with multiple face images in passport size on it\newline{}2.	Printed using DNP DS820\newline{}3.	Scanned with Epson XP-860 at 300 dpi\newline{}4.	Crop individual face images to 420x540 \newline{}5. Compressed using JPEG2000 (max file size = 15kb, RGB\_24\_BIT) \bigstrut\\
    \hline 
    F03 & 1.	Printed with printer Dmb DS-RX1HS at 300x300 dpi, matte\newline{}2.	Scanned using Epson V600 at 300 dpi\newline{}3.	Cropped to 413 x 531 pixels\newline{}4.	Compressed using JPEG2000, max file size=15kb, RGB\_24\_BIT & F22 & 1.	Created A4 PDFs with multiple face images in passport size on it\newline{}2.	Printed using DNP DS820\newline{}3.	Scan with Canon TS8251 at 300 dpi\newline{}4. Crop individual face images to 420x5404 \newline{}5. Compressed using JPEG2000 (max file size = 15kb, RGB\_24\_BIT)\bigstrut\\
    \hline
    F04 & 1.	Printed with printer Dmb DS-RX1HS at 300x300 dpi, matte\newline{}2.	Scanned using ID Document Scanner (Idemia) \newline{}3.	Retrieved (compressed) image & F23 & 1. Created A4 PDFs with multiple face images in passport size on it\newline{}2. Printed using Epson XP-860\newline{}3. Scanned with Epson XP-860 at 300 dpi\newline{}4. Cropped individual face images to 420x5404 \newline{}5. Compressed using JPEG2000 (max file size = 15kb, RGB\_24\_BIT) \bigstrut\\
    \hline 
    F05 & 1.	Printed with printer Dmb DS-RX1HS at 300x600 dpi, matte\newline{}2.	Scanned using ID Document Scanner (Idemia) \newline{}3.	Retrieved (compressed) image & F24 & 1. Created A4 PDFs with multiple face images in passport size on it\newline{}2. Printed using Epson XP-860\newline{}3. Scanned with Canon TS8251 at 300 dpi\newline{}4. Cropped individual face images to 420x5404 \newline{} 5. Compressed using JPEG2000 (max file size = 15kb, RGB\_24\_BIT) \bigstrut\\    
    \hline 
    F11 & 1.	Created A4 PDFs with multiple face images in passport size on it\newline{}2.	Printed PDF at professional photo laboratory of Fotogena GmbH \newline{} to A4 image paper\newline{}3.	Scanned with Canon Lide 220 at 300 dpi, sharpness filter \newline{} + denoise low as .png\newline{}4.	Recreated the images from the scanned A4 to independent images \newline{}5. Compressed using JPEG2000 (max file size = 15kb, RGB\_24\_BIT) & F25 & 1. Printed the face images using DNP DS820\newline{}2. Scanned using ID Document Scanner (Idemia) \newline{}3. Retrieved (compressed) image JPEG2000,max file size=15kb \bigstrut\\
    \hline
    F12 & 1.	Created A4 PDFs with multiple face images in passport size on it\newline{}2.	Printed PDF at professional photo laboratory of Fotogena GmbH \newline{} to A4 image paper\newline{}3.	Scanned with Kyocera Ecosys M6035cidn at 300 dpi, sharpness +1, \newline{} no other optimizations as tiff\newline{}4.	Recreated the images from the scanned A4\newline{}5. Compressed using JPEG2000 (max file size = 15kb, RGB\_24\_BIT) & F26 & 1. Printed the face images using Epson XP-860\newline{}2. Scanned using ID Document Scanner (Idemia) \newline{}3. Retrieved (compressed) image JPEG2000,max file size=15kb \bigstrut\\
    \hline 
    \end{tabular}%
}
  \label{tab:print-scan-details}%
\end{table}%

\begin{table*}[htbp]
  \centering
  \caption{Subset EER deviation w.r.t. the overall set of digital images with morphing factor 0.3.}
  \resizebox{0.9\textwidth}{!}{
    \begin{tabular}{|l|l|r|c|r|r|r|r|r|r|}
    \hline
    \multicolumn{1}{|c|}{\textbf{Attribute}} & \multicolumn{1}{c|}{\textbf{Subset}} & \multicolumn{1}{c|}{\textbf{$\overline{dev_s}$}} & \multirow{32}[64]{*}{} & \multicolumn{1}{c|}{\textbf{BSIF}} & \multicolumn{1}{c|}{\textbf{DFR}} & \multicolumn{1}{c|}{\textbf{MBLBP}} & \multicolumn{1}{c|}{\textbf{WL}} & \multicolumn{1}{c|}{\textbf{DR}} & \multicolumn{1}{c|}{\textbf{FaDe}} \\
\cline{1-3}\cline{5-10}    Ethnicity & Middle Eastern & \cellcolor[rgb]{ .388.  .745.  .482}-31.36\% &       & \cellcolor[rgb]{ .914.  .894.  .51}-4.03\% & \cellcolor[rgb]{ .388.  .745.  .482}-100.00\% & \cellcolor[rgb]{ .675.  .827.  .498}-13.43\% & \cellcolor[rgb]{ .388.  .745.  .482}-30.36\% & \cellcolor[rgb]{ 1.  .902.  .514}0.34\% & \cellcolor[rgb]{ .388.  .745.  .482}-40.69\% \\
\cline{1-3}\cline{5-10}    Ethnicity & Indian-Asian & \cellcolor[rgb]{ .388.  .745.  .482}-30.76\% &       & \cellcolor[rgb]{ .486.  .773.  .486}-20.87\% & \cellcolor[rgb]{ .388.  .745.  .482}-63.27\% & \cellcolor[rgb]{ .388.  .745.  .482}-26.96\% & \cellcolor[rgb]{ .388.  .745.  .482}-26.41\% & \cellcolor[rgb]{ .851.  .878.  .506}-6.39\% & \cellcolor[rgb]{ .388.  .745.  .482}-40.69\% \\
\cline{1-3}\cline{5-10}    Partner & HDA   & \cellcolor[rgb]{ .388.  .745.  .482}-27.13\% &       & \cellcolor[rgb]{ .851.  .878.  .506}-6.53\% & \cellcolor[rgb]{ .388.  .745.  .482}-90.31\% & \cellcolor[rgb]{ .388.  .745.  .482}-46.02\% & \cellcolor[rgb]{ .82.  .867.  .506}-7.73\% & \cellcolor[rgb]{ .941.  .902.  .514}-2.83\% & \cellcolor[rgb]{ .776.  .855.  .502}-9.37\% \\
\cline{1-3}\cline{5-10}    Partner & NTNU  & \cellcolor[rgb]{ .475.  .769.  .486}-21.30\% &       & \cellcolor[rgb]{ .463.  .765.  .486}-21.77\% & \cellcolor[rgb]{ .388.  .745.  .482}-54.08\% & \cellcolor[rgb]{ .486.  .773.  .486}-20.81\% & \cellcolor[rgb]{ .918.  .898.  .51}-3.86\% & \cellcolor[rgb]{ .996.  .8.  .498}5.00\% & \cellcolor[rgb]{ .388.  .745.  .482}-32.27\% \\
\cline{1-3}\cline{5-10}    Age   & 18..35 & \cellcolor[rgb]{ .627.  .812.  .494}-15.30\% &       & \cellcolor[rgb]{ .906.  .894.  .51}-4.26\% & \cellcolor[rgb]{ .388.  .745.  .482}-56.63\% & \cellcolor[rgb]{ .733.  .843.  .502}-11.15\% & \cellcolor[rgb]{ .855.  .878.  .506}-6.30\% & \cellcolor[rgb]{ 1.  .871.  .51}1.73\% & \cellcolor[rgb]{ .631.  .812.  .494}-15.18\% \\
\cline{1-3}\cline{5-10}    Ethnicity & East-Asian & \cellcolor[rgb]{ .675.  .827.  .498}-13.41\% &       & \cellcolor[rgb]{ .996.  .792.  .494}5.34\% & \cellcolor[rgb]{ .388.  .745.  .482}-100.00\% & \cellcolor[rgb]{ .686.  .827.  .498}-13.02\% & \cellcolor[rgb]{ .98.  .506.  .439}18.59\% & \cellcolor[rgb]{ .988.  .694.  .475}9.90\% & \cellcolor[rgb]{ .98.  .914.  .514}-1.30\% \\
\cline{1-3}\cline{5-10}    Traits & None  & \cellcolor[rgb]{ .769.  .855.  .502}-9.65\% &       & \cellcolor[rgb]{ .89.  .89.  .51}-4.87\% & \cellcolor[rgb]{ .388.  .745.  .482}-31.63\% & \cellcolor[rgb]{ .976.  .914.  .514}-1.58\% & \cellcolor[rgb]{ .906.  .894.  .51}-4.29\% & \cellcolor[rgb]{ 1.  .898.  .514}0.61\% & \cellcolor[rgb]{ .608.  .808.  .494}-16.13\% \\
\cline{1-3}\cline{5-10}    Morph quality & Low   & \cellcolor[rgb]{ .863.  .882.  .51}-5.96\% &       & \cellcolor[rgb]{ .996.  .839.  .502}3.17\% & \cellcolor[rgb]{ .486.  .773.  .486}-20.92\% & \cellcolor[rgb]{ .922.  .898.  .51}-3.61\% & \cellcolor[rgb]{ .949.  .906.  .514}-2.56\% & \cellcolor[rgb]{ 1.  .902.  .514}0.40\% & \cellcolor[rgb]{ .706.  .835.  .498}-12.22\% \\
\cline{1-3}\cline{5-10}    Morph. algorithm & C07   & \cellcolor[rgb]{ .871.  .882.  .51}-5.76\% &       & \cellcolor[rgb]{ 1.  .871.  .51}1.77\% & \cellcolor[rgb]{ .588.  .8.  .49}-16.84\% & \cellcolor[rgb]{ .906.  .894.  .51}-4.34\% & \cellcolor[rgb]{ .996.  .918.  .514}-0.67\% & \cellcolor[rgb]{ .98.  .914.  .514}-1.33\% & \cellcolor[rgb]{ .682.  .827.  .498}-13.17\% \\
\cline{1-3}\cline{5-10}    Morph. algorithm & C06   & \cellcolor[rgb]{ .878.  .886.  .51}-5.32\% &       & \cellcolor[rgb]{ .824.  .871.  .506}-7.60\% & \cellcolor[rgb]{ 1.  .886.  .514}1.02\% & \cellcolor[rgb]{ .761.  .851.  .502}-9.98\% & \cellcolor[rgb]{ .694.  .831.  .498}-12.69\% & \cellcolor[rgb]{ .969.  .91.  .514}-1.83\% & \cellcolor[rgb]{ .992.  .918.  .514}-0.83\% \\
\cline{1-3}\cline{5-10}    Post-processing & PM03  & \cellcolor[rgb]{ .898.  .89.  .51}-4.58\% &       & \cellcolor[rgb]{ .973.  .412.  .42}50.72\% & \cellcolor[rgb]{ .388.  .745.  .482}-27.04\% & \cellcolor[rgb]{ .984.  .914.  .514}-1.24\% & \cellcolor[rgb]{ .988.  .651.  .467}11.90\% & \cellcolor[rgb]{ .784.  .859.  .502}-9.03\% & \cellcolor[rgb]{ .388.  .745.  .482}-52.79\% \\
\cline{1-3}\cline{5-10}    Morph. algorithm & C03   & \cellcolor[rgb]{ .929.  .898.  .51}-3.36\% &       & \cellcolor[rgb]{ .988.  .635.  .463}12.69\% & \cellcolor[rgb]{ .486.  .773.  .486}-20.92\% & \cellcolor[rgb]{ .89.  .89.  .51}-4.91\% & \cellcolor[rgb]{ .996.  .843.  .506}3.07\% & \cellcolor[rgb]{ 1.  .906.  .518}0.23\% & \cellcolor[rgb]{ .753.  .851.  .502}-10.32\% \\
\cline{1-3}\cline{5-10}    Ethnicity & European/Amer. & \cellcolor[rgb]{ .949.  .906.  .514}-2.65\% &       & \cellcolor[rgb]{ 1.  .922.  .518}-0.58\% & \cellcolor[rgb]{ .992.  .722.  .482}8.67\% & \cellcolor[rgb]{ .82.  .871.  .506}-7.63\% & \cellcolor[rgb]{ 1.  .863.  .506}2.19\% & \cellcolor[rgb]{ 1.  .878.  .51}1.50\% & \cellcolor[rgb]{ .506.  .776.  .486}-20.05\% \\
\cline{1-3}\cline{5-10}    Post-processing & PM01  & \cellcolor[rgb]{ .969.  .91.  .514}-1.82\% &       & \cellcolor[rgb]{ .996.  .918.  .514}-0.78\% & \cellcolor[rgb]{ .769.  .855.  .502}-9.69\% & \cellcolor[rgb]{ .976.  .431.  .424}22.11\% & \cellcolor[rgb]{ .992.  .851.  .486}7.45\% & \cellcolor[rgb]{ .992.  .918.  .514}-0.95\% & \cellcolor[rgb]{ .388.  .745.  .482}-29.06\% \\
\cline{1-3}\cline{5-10}    Gender & Male  & \cellcolor[rgb]{ 1.  .922.  .518}-0.63\% &       & \cellcolor[rgb]{ .949.  .906.  .514}-2.54\% & \cellcolor[rgb]{ .996.  .788.  .494}5.61\% & \cellcolor[rgb]{ .953.  .906.  .514}-2.41\% & \cellcolor[rgb]{ .996.  .82.  .498}4.17\% & \cellcolor[rgb]{ 1.  .859.  .506}2.28\% & \cellcolor[rgb]{ .737.  .843.  .502}-10.91\% \\
\cline{1-3}\cline{5-10}    Post-processing & Automatic & \cellcolor[rgb]{ 1.  .922.  .518}-0.60\% &       & \cellcolor[rgb]{ 1.  .91.  .518}0.06\% & \cellcolor[rgb]{ 1.  .922.  .518}-0.51\% & \cellcolor[rgb]{ .996.  .918.  .514}-0.76\% & \cellcolor[rgb]{ .973.  .914.  .514}-1.64\% & \cellcolor[rgb]{ 1.  .898.  .514}0.44\% & \cellcolor[rgb]{ .984.  .914.  .514}-1.19\% \\
\cline{1-3}\cline{5-10}    Morph. algorithm & C05   & \cellcolor[rgb]{ 1.  .918.  .518}-0.33\% &       & \cellcolor[rgb]{ 1.  .922.  .518}-0.54\% & \cellcolor[rgb]{ 1.  .91.  .518}0.00\% & \cellcolor[rgb]{ 1.  .914.  .518}-0.13\% & \cellcolor[rgb]{ 1.  .867.  .51}2.07\% & \cellcolor[rgb]{ .996.  .918.  .514}-0.76\% & \cellcolor[rgb]{ .949.  .906.  .514}-2.61\% \\
\cline{1-3}\cline{5-10}    Post-processing & Manual & \cellcolor[rgb]{ 1.  .875.  .51}1.65\% &       & \cellcolor[rgb]{ 1.  .918.  .518}-0.45\% & \cellcolor[rgb]{ 1.  .91.  .518}0.00\% & \cellcolor[rgb]{ 1.  .867.  .51}2.06\% & \cellcolor[rgb]{ .996.  .82.  .498}4.08\% & \cellcolor[rgb]{ .988.  .918.  .514}-1.10\% & \cellcolor[rgb]{ .996.  .792.  .494}5.34\% \\
\cline{1-3}\cline{5-10}    Post-processing & PM06  & \cellcolor[rgb]{ .996.  .839.  .502}3.34\% &       & \cellcolor[rgb]{ .388.  .745.  .482}-42.32\% & \cellcolor[rgb]{ .973.  .412.  .42}103.06\% & \cellcolor[rgb]{ .388.  .745.  .482}-39.66\% & \cellcolor[rgb]{ .412.  .851.  .482}-23.79\% & \cellcolor[rgb]{ .937.  .902.  .514}-3.00\% & \cellcolor[rgb]{ .973.  .412.  .42}25.74\% \\
\cline{1-3}\cline{5-10}    Ethnicity & African & \cellcolor[rgb]{ .996.  .816.  .498}4.36\% &       & \cellcolor[rgb]{ .992.  .725.  .482}8.46\% & \cellcolor[rgb]{ .984.  .569.  .451}15.82\% & \cellcolor[rgb]{ 1.  .886.  .514}1.11\% & \cellcolor[rgb]{ .867.  .882.  .51}-5.84\% & \cellcolor[rgb]{ .886.  .886.  .51}-5.13\% & \cellcolor[rgb]{ .988.  .655.  .467}11.74\% \\
\cline{1-3}\cline{5-10}    Post-processing & PM02  & \cellcolor[rgb]{ .992.  .769.  .49}6.50\% &       & \cellcolor[rgb]{ .839.  .875.  .506}-6.89\% & \cellcolor[rgb]{ .875.  .882.  .51}-5.61\% & \cellcolor[rgb]{ .976.  .427.  .424}22.27\% & \cellcolor[rgb]{ .988.  .686.  .475}10.34\% & \cellcolor[rgb]{ .996.  .796.  .494}5.27\% & \cellcolor[rgb]{ .984.  .616.  .459}13.64\% \\
\cline{1-3}\cline{5-10}    Morph. algorithm & C02   & \cellcolor[rgb]{ .992.  .761.  .49}6.84\% &       & \cellcolor[rgb]{ .867.  .882.  .51}-5.82\% & \cellcolor[rgb]{ .835.  .871.  .506}-7.14\% & \cellcolor[rgb]{ .976.  .443.  .427}21.57\% & \cellcolor[rgb]{ .988.  .682.  .475}10.50\% & \cellcolor[rgb]{ .992.  .718.  .478}8.90\% & \cellcolor[rgb]{ .984.  .627.  .463}13.05\% \\
\cline{1-3}\cline{5-10}    Gender & Female & \cellcolor[rgb]{ .988.  .671.  .471}11.01\% &       & \cellcolor[rgb]{ 1.  .875.  .51}1.68\% & \cellcolor[rgb]{ .973.  .412.  .42}71.43\% & \cellcolor[rgb]{ .996.  .788.  .494}5.64\% & \cellcolor[rgb]{ .914.  .894.  .51}-3.95\% & \cellcolor[rgb]{ .894.  .89.  .51}-4.81\% & \cellcolor[rgb]{ .914.  .894.  .51}-3.91\% \\
\cline{1-3}\cline{5-10}    Morph. algorithm & C01   & \cellcolor[rgb]{ .984.  .624.  .463}13.24\% &       & \cellcolor[rgb]{ 1.  .898.  .514}0.60\% & \cellcolor[rgb]{ .973.  .412.  .42}62.24\% & \cellcolor[rgb]{ 1.  .875.  .51}1.55\% & \cellcolor[rgb]{ 1.  .882.  .51}1.34\% & \cellcolor[rgb]{ .984.  .918.  .514}-1.14\% & \cellcolor[rgb]{ .984.  .588.  .455}14.83\% \\
\cline{1-3}\cline{5-10}    Age   & 36..55 & \cellcolor[rgb]{ .984.  .612.  .459}13.84\% &       & \cellcolor[rgb]{ .988.  .694.  .475}9.93\% & \cellcolor[rgb]{ .973.  .412.  .42}53.06\% & \cellcolor[rgb]{ .976.  .482.  .435}19.70\% & \cellcolor[rgb]{ .996.  .784.  .494}5.81\% & \cellcolor[rgb]{ .949.  .906.  .514}-2.64\% & \cellcolor[rgb]{ .941.  .902.  .514}-2.85\% \\
\cline{1-3}\cline{5-10}    Post-processing & PM05  & \cellcolor[rgb]{ .984.  .6.  .459}14.38\% &       & \cellcolor[rgb]{ .965.  .91.  .514}-2.00\% & \cellcolor[rgb]{ .973.  .412.  .42}69.39\% & \cellcolor[rgb]{ .969.  .91.  .514}-1.87\% & \cellcolor[rgb]{ .992.  .706.  .478}9.46\% & \cellcolor[rgb]{ .984.  .918.  .514}-1.18\% & \cellcolor[rgb]{ .988.  .639.  .467}12.46\% \\
\cline{1-3}\cline{5-10}    Traits & Freckels & \cellcolor[rgb]{ .984.  .596.  .455}14.52\% &       & \cellcolor[rgb]{ .988.  .651.  .467}12.00\% & \cellcolor[rgb]{ .973.  .412.  .42}53.06\% & \cellcolor[rgb]{ .969.  .91.  .514}-1.84\% & \cellcolor[rgb]{ .992.  .722.  .482}8.67\% & \cellcolor[rgb]{ 1.  .851.  .506}2.66\% & \cellcolor[rgb]{ .988.  .639.  .463}12.57\% \\
\cline{1-3}\cline{5-10}    Age   & 56..75 & \cellcolor[rgb]{ .973.  .412.  .42}32.74\% &       & \cellcolor[rgb]{ .984.  .569.  .451}15.74\% & \cellcolor[rgb]{ .973.  .412.  .42}89.29\% & \cellcolor[rgb]{ .973.  .412.  .42}31.83\% & \cellcolor[rgb]{ .973.  .412.  .42}36.26\% & \cellcolor[rgb]{ 1.  .922.  .518}-0.65\% & \cellcolor[rgb]{ .973.  .412.  .42}23.96\% \\
\cline{1-3}\cline{5-10}    Partner & UTW   & \cellcolor[rgb]{ .973.  .412.  .42}32.79\% &       & \cellcolor[rgb]{ .992.  .765.  .49}6.66\% & \cellcolor[rgb]{ .973.  .412.  .42}142.35\% & \cellcolor[rgb]{ .992.  .757.  .486}7.03\% & \cellcolor[rgb]{ .988.  .69.  .475}10.10\% & \cellcolor[rgb]{ .929.  .898.  .51}-3.35\% & \cellcolor[rgb]{ .973.  .412.  .42}33.93\% \\
\cline{1-3}\cline{5-10}    Morph quality & High  & \cellcolor[rgb]{ .973.  .412.  .42}41.48\% &       & \cellcolor[rgb]{ .737.  .843.  .502}-10.96\% & \cellcolor[rgb]{ .973.  .412.  .42}204.59\% & \cellcolor[rgb]{ .988.  .659.  .467}11.69\% & \cellcolor[rgb]{ .988.  .675.  .471}10.80\% & \cellcolor[rgb]{ .976.  .914.  .514}-1.52\% & \cellcolor[rgb]{ .973.  .412.  .42}34.28\% \\
\cline{1-3}\cline{5-10}    Traits & Moles & \cellcolor[rgb]{ .973.  .412.  .42}44.52\% &       & \cellcolor[rgb]{ .996.  .8.  .494}5.15\% & \cellcolor[rgb]{ .973.  .412.  .42}188.27\% & \cellcolor[rgb]{ .973.  .412.  .42}43.87\% & \cellcolor[rgb]{ 1.  .863.  .506}2.22\% & \cellcolor[rgb]{ .992.  .714.  .478}8.99\% & \cellcolor[rgb]{ .98.  .506.  .439}18.62\% \\
    \hline
    \end{tabular}%
    }
  \label{tab:subsetRes_D0.3}%
\end{table*}%

\end{appendices}

\end{document}